\RequirePackage{fix-cm}
\documentclass[]{template}

\usepackage{iftex}
\ifPDFTeX\usepackage[utf8]{inputenc}\fi
\iftutex\fi
\usepackage{amsmath,amssymb,amsfonts}
\usepackage{url}
\usepackage{xspace}
\usepackage{listings}
\usepackage{enumitem}
\usepackage{wrapfig}
\usepackage{colortbl}
\usepackage{pifont}
\usepackage{makecell}
\usepackage{array}
\usepackage[numbers,sort&compress]{natbib}

\newcommand{\ours}{AutoPILOT\xspace}
\newcommand{\bench}{AFK-Bench\xspace}
\newcommand{\vista}{VISTA\xspace}
\newcommand{\tool}[1]{\texttt{#1}}
\newcommand{\gpt}{GPT-6.1 Sol\xspace}
\newcommand{\opus}{Claude Opus 5.5\xspace}
\newcommand{\hrt}{hostile real-time\xspace}
\newcommand{\brt}{benign real-time\xspace}
\newcommand{\tb}{turn-based\xspace}
\newcommand{\Hrt}{Hostile real-time\xspace}
\newcommand{\Brt}{Benign real-time\xspace}
\newcommand{\Tb}{Turn-based\xspace}
\newcommand{\prog}[1]{{\scriptsize\color{gray}.#1}}

\newcommand{\best}[1]{\textbf{#1}}
\definecolor{apblue}{HTML}{2A6FB0}
\definecolor{apbg}{HTML}{E6EFF8}
\definecolor{vorange}{HTML}{D4762C}
\newcommand{\oursrow}{\rowcolor{apbg}}
\newcommand{\yes}{{\color{apblue}\ding{51}}}
\newcommand{\no}{{\color{gray}\ding{55}}}
\newcommand{\partly}{$\sim$}
\newcommand{\na}{--}
\makeatletter
\DeclareRobustCommand\onedot{\futurelet\@let@token\@onedot}
\def\@onedot{\ifx\@let@token.\else.\null\fi\xspace}
\makeatother
\def\eg{\emph{e.g}\onedot}

 \def\vs{\emph{vs}\onedot}
 \def\etal{\emph{et al}\onedot}

\definecolor{airforceblue}{rgb}{0.36, 0.54, 0.66}
\definecolor{bluegray}{rgb}{0.4, 0.6, 0.8}
\definecolor{bleudefrance}{rgb}{0.19, 0.55, 0.91}
\usepackage[breaklinks]{hyperref}
\hypersetup{colorlinks,linkcolor={airforceblue},citecolor={bleudefrance},urlcolor={bluegray}}
\usepackage[capitalize,noabbrev,nameinlink]{cleveref}
\crefname{section}{Sec.}{Secs.}
\Crefname{section}{Section}{Sections}
\crefname{figure}{Fig.}{Figs.}
\crefname{table}{Tab.}{Tabs.}
\Crefname{table}{Table}{Tables}
\crefname{equation}{Eq.}{Eqs.}
\crefname{appendix}{App.}{Apps.}

\renewenvironment{abstract}{}{}
\newcounter{mainpages}
\newcommand{\maketitlesupplementary}{\setcounter{page}{\value{mainpages}}\beginappendix}

\usepackage{needspace}

\title{AutoPILOT: A Visual Agent Harness \\for Real-Time Gameplay}
\renewcommand{\shorttitle}{AutoPILOT: A Visual Agent Harness for Real-Time Gameplay}

\author[]{%
Mark Zhao$^{1,2*\dagger}$,\hspace{0.4em}
Pengcheng Wang$^{1*}$,\hspace{0.4em}
Rui Yang$^{1*}$,\hspace{0.4em}
Jiarui Yao$^{1}$,\hspace{0.4em}
Xirui Jiang$^{1}$,\hspace{0.4em}
Yuxi Chen$^{1}$,\hspace{0.4em}
Bingxuan Li$^{1}$\\
Hanyang Chen$^{1}$,\hspace{0.4em}
Ke Yang$^{1}$,\hspace{0.4em}
Manling Li$^{3}$,\hspace{0.4em}
Tong Zhang$^{1}$,\hspace{0.4em}
Huan Zhang$^{1}$%
}

\affiliation{%
$^1$University of Illinois Urbana-Champaign,\hspace{6pt}
$^2$Georgia Institute of Technology,\hspace{6pt}
$^3$Northwestern University\\[2pt]
{\urlstyle{rm}\url{https://autopilot-harness.github.io/}}%
}

\setfirstpagefooter{%
$^*$Equal contribution. $^\dagger$Work done as an intern at UIUC. Emails: \texttt{zzhao661@gatech.edu}, \texttt{ry21@illinois.edu}%
}

\begin{document}

\setteaser{%
\centering
\vskip 3mm
\includegraphics[width=\linewidth]{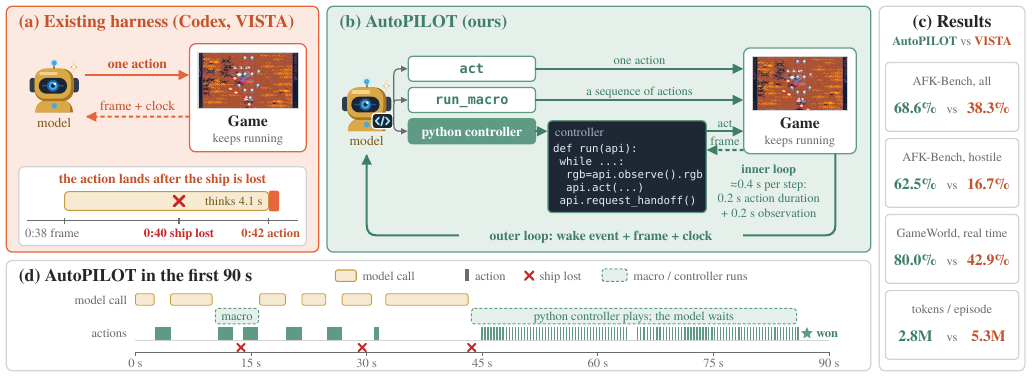}\par
\vskip 0.5mm
\captionof{figure}{\textbf{Existing harnesses \vs \ours on \emph{Praxis Fighter X} in real time.}
  (a)~Existing harness acts once per model call, so the action lands after the ship is lost. Given the same restarts as \ours, it keeps losing.
  (b)~\ours can also start a macro or a model-written controller that keeps playing until a wake event.
  (c)~Real-time results with \gpt on \bench and GameWorld~\cite{ouyang2026gameworld}.
  (d)~The first 90\,s of \ours: after three losses, a controller takes over and wins at 1:26.}
\label{fig:teaser}
}
\paperabstract{\begin{abstract}
Multimodal foundation models can play video games well when the game waits for them, but often fail when it keeps running. We find that much of this gap lies in the harness, which decides what the model sees, what it remembers, and how its outputs become actions over time. Existing harnesses such as \vista improve the first two, yet turn each model call into at most one action, so the agent acts only as often as its model thinks. We introduce \ours, which changes the third: in each model call, the model can send a primitive action, start an open-loop macro, or write a closed-loop controller that keeps playing from the screen without further model calls until it hands control back. To evaluate agents in real-time games, we build \bench, 120 diverse browser games grouped by how they respond to inaction and playable in real time or with the clock paused. With \gpt under the same budgets, \ours reaches 68.6\% real-time success against 38.3\% for \vista, and on GameWorld it keeps 80.0\% in real time while \vista drops from 84.1\% to 42.9\%. Its gains hold across closed and open-source models, and it reaches \vista's final success rate at a fifth of its API cost. Our results suggest that how a harness turns model outputs into actions over time is a central yet underexplored part of agent design, and we hope \ours and \bench will serve as a foundation for agents in a world that does not wait.
\end{abstract}
}

\maketitle

\section{Introduction}
\label{sec:intro}

\begin{table*}[t]
  \centering
  \footnotesize
  \setlength{\tabcolsep}{3pt}
  \renewcommand{\arraystretch}{0.92}
  \resizebox{\linewidth}{!}{\begin{tabular}{@{}llccccclr@{}}
    \toprule
    Method & Observation & \makecell{No game-specific\\training} & \makecell{One harness\\for many games} & \makecell{Real-time\\evaluation} & \makecell{Macro\\actions} & \makecell{Model-written\\controller} & Env.\ type & \# Envs \\
    \midrule
    Voyager~\cite{wang2024voyager} & game state & \yes & \no & \na & \yes & \yes & Minecraft & 1 \\
    Code as Policies~\cite{liang2023codeaspolicies} & perception API & \yes & \no & \na & \yes & \yes & robotics & 3 \\
    Cradle~\cite{tan2025cradle} & screen & \yes & \yes & \no & \yes & \no & PC games, apps & 10 \\
    Game-TARS~\cite{wang2025gametars} & screen & \no & \yes & \no & \no & \no & video games & 7 \\
    HLA~\cite{liu2024hla} & text & \yes & \no & \yes & \yes & \no & Overcooked & 1 \\
    DPT-Agent~\cite{zhang2025dptagent} & text & \yes & \no & \yes & \yes & \partly & Overcooked & 1 \\
    AgileThinker~\cite{wen2026realtime} & text & \yes & \yes & \yes & \no & \no & grid games & 3 \\
    Lumine~\cite{tan2025lumine} & screen & \no & \yes & \yes & \yes & \no & 3D action games & 4 \\
    TORSI~\cite{pan2026worldwontwait} & text & \yes & \no & \yes & \partly & \no & games & 6 \\
    Realtime-Bench~\cite{tang2026realtimebench} & screen & \yes & \yes & \yes & \no & \no & web, games & 50 \\
    AsyncLLM~\cite{yakushev2026async} & frames, text & \yes & \yes & \yes & \no & \no & QA, video, ViZDoom & 6 \\
    AutoHarness~\cite{lou2026autoharness} & text & \yes & \no & \no & \no & \yes & text games & 145 \\
    \vista~\cite{han2026vista} & screen & \yes & \yes & \no & \no & \no & games, puzzles & 70 \\
    \midrule
    \oursrow \ours (ours) & screen & \yes & \yes & \yes & \yes & \yes & browser games & 154 \\
    \bottomrule
  \end{tabular}}
  \vspace{-0.5em}
  \caption{\textbf{Comparison with game agents and harnesses.} \partly~partly (DPT-Agent uses rules for a hand-built state machine, and TORSI makes multi-step plans in 2 of 6 games). \na~not reported. Real-time: the game runs during model calls. Macro actions: several actions per model call. \# Envs: distinct games, apps or benchmarks (ours: 120 in \bench, 34 in GameWorld).}
  \label{tab:related}
  \vspace{-1em}
\end{table*}

Multimodal foundation models can now play video games~\cite{simateam2025sima2,tan2025cradle,wang2025gametars,tan2025lumine}, operate graphical interfaces~\cite{cheng2024seeclick,qin2025uitars,xie2024osworld,lu2026qwencua,yang2026guilibra}, and solve visual puzzles~\cite{arcagi3,lehmann2026tycho} from pixels.
Their success depends heavily on the \emph{harness}, the system around the model~\cite{weng2026harness}: with the model unchanged, a better harness can raise success substantially~\cite{yang2024sweagent}, and \vista~\cite{han2026vista} took an off-the-shelf model to a perfect score on ARC-AGI-3~\cite{arcagi3}.
A harness decides what the model sees and remembers, and how its outputs become actions over time.
Prior work has mostly refined the first two~\cite{yang2024sweagent,tan2025cradle,hu2025lmgamebench,zhou2026openvlharness,han2026vista}.
The third is usually left implicit, following the standard \emph{single-step} agent loop~\cite{yao2023react} in which one model call yields at most one action and the environment is assumed to wait for it, as in ARC-AGI-3 or game benchmarks with settings that pause during inference~\cite{ying2026aigamestore,zhang2025videogamebench}.

When the world keeps running, this default becomes a bottleneck: the agent can act only as often as its model thinks, and on outdated frames (\cref{fig:teaser}a).
Letting the game clock run lowers \vista's success rate on GameWorld~\cite{ouyang2026gameworld} from 84.1\% to 42.9\%, a gap also reported by Realtime-Bench~\cite{tang2026realtimebench} and for text-state agents~\cite{pan2026worldwontwait}.
Improving what the model sees and remembers does not remove it, as \vista performs on par with the plain Codex harness in real time despite its visual memory and inspection tools (38.3\% \vs 38.9\%).
Nor does more reasoning, which adds 4--7 points to \vista at high effort.
Much of the bottleneck lies in how the harness turns model outputs into actions.

We introduce \textbf{\ours}, a visual agent harness that changes the third aspect by letting one model call cover many environment steps.
Like a pilot with an autopilot, the model delegates routine control and takes over when called back.
Each call chooses one of three execution modes (\cref{fig:teaser}b).
(1) A \emph{primitive action} is sent directly to the game.
(2) An \emph{open-loop macro} is a sequence of actions written by the model.
(3) A \emph{closed-loop controller} is a short Python program that the model writes at run time.
It reads the screen and keeps acting without further model calls.
Control returns to the model on a \emph{wake event}: the macro or controller finishes, requests a handoff, or meets a screen it was not written for.
Both the model and its controllers see screenshots, not the game state.

To evaluate agents in real time, we introduce \textbf{\bench} (\emph{Away From Keyboard}), 120 browser games evenly divided into three types by what happens while the agent is idle.
In \emph{\hrt environments}, inaction risks failure.
\emph{\Brt environments} keep changing but do not punish inaction.
\emph{\Tb environments} advance only when the agent acts.
A lost game restarts within fixed budgets of model calls, game time and environment steps.
Every game can be played in real time or paused, and the difference measures what real time costs the agent.

With the same model (\gpt) and budgets, \ours succeeds in 68.6\% of real-time episodes on \bench, against 38.3\% for \vista, and in 62.5\% \vs 16.7\% of \hrt ones.
Running the clock costs \ours 4.5 points on \bench and 2.9 on GameWorld, against 17.8 and 41.2 for \vista, and \ours leads even with the clock paused (73.1\% \vs 56.1\%), because one model call can execute a whole plan.
The gains hold with \opus, three open-source models and one more closed model, and \ours uses 47\% fewer tokens and reaches \vista's final success rate at 0.2$\times$ its API cost (\cref{fig:curves}d).

Our contributions are:
\begin{itemize}[leftmargin=*,itemsep=1pt,topsep=2pt]
  \item \textbf{Acting over time as harness design.} We treat how model outputs become actions over time as a design choice of the harness, and propose \ours, in which one model call can start a model-written macro or controller (\cref{sec:method}).
  \item \textbf{\bench.} We introduce \bench, 120 browser games measuring what real time costs an agent (\cref{sec:bench}).
  \item \textbf{Detailed analysis.} \ours outperforms \vista in real time at lower cost and, on \bench, even with the clock paused. We analyze how its components, budgets, reasoning effort and generation speed matter (\cref{sec:exp}).
\end{itemize}

\begin{figure*}[t]
\centering
\begin{minipage}[t]{0.29\textwidth}
\textbf{\small (a) Open-loop macro} \hfill {\scriptsize\color{gray}Trichromat}
\begin{lstlisting}[language=Python,basicstyle=\ttfamily\fontsize{8}{12.6}\selectfont]
# model, 1:36: "I'll send
# the nine moves directly."
# 9 keys, each held 0.15 s
run_macro([
  'ArrowDown', 'ArrowLeft',
  'ArrowLeft', 'ArrowDown',
  'ArrowRight','ArrowRight',
  'ArrowLeft', 'ArrowDown',
  'ArrowRight'])
# worker starts one every
# 0.36 s without reading the
# screen, then wakes the
# model: "macro done"
\end{lstlisting}
\end{minipage}\hfill
\begin{minipage}[t]{0.69\textwidth}
\textbf{\small (b) Closed-loop controller} \hfill {\scriptsize\color{gray}Praxis Fighter X}
\begin{lstlisting}[language=Python,basicstyle=\ttfamily\fontsize{6.6}{7.6}\selectfont,breaklines=true,breakindent=24pt,postbreak=\mbox{$\hookrightarrow$\space}]
def run(api):
 import numpy as np
 for i in range(240):
  o=api.observe(); a=o.rgb
  # Blue hull pixels near the bottom identify current horizontal position.
  m=(a[560:680,:,2]>100)&(a[560:680,:,0]<130)&(a[560:680,:,1]>70)
  yy,xx=np.where(m); x=float(np.median(xx)) if len(xx)>10 else 640
  # Bright magenta/red projectiles, excluding the side HUD.
  b=(a[:,:,0]>235)&(a[:,:,1]<115)&(a[:,:,2]>35)
  b[:300]=False; b[:, :100]=False; b[:,1180:]=False
  by,bx=np.where(b)
  near=(abs(bx-x)<150)&(by>390)
  if np.any(near):
   threat=float(np.mean(bx[near]))
   direction='ArrowRight' if threat<x else 'ArrowLeft'
  else:
   direction='ArrowRight' if (i//22)%2==0 else 'ArrowLeft'
  if x<220: direction='ArrowRight'
  if x>1060: direction='ArrowLeft'
  api.act({'action':'press_keys','keys':['x',direction],'duration':0.18})
 api.request_handoff('Short firing/dodging trial finished; inspect score and lives.')
\end{lstlisting}
\end{minipage}
\vspace{-0.6em}
\caption{\textbf{Examples of a macro and a controller in real-time episodes of \gpt.} (a)~A macro in the \tb game \emph{Trichromat}, compacted from the action log. Key presses start about every 0.36\,s: each key is held for 0.15\,s, a frame is captured after release (about 0.14\,s), and the next key follows (about 0.06\,s). (b)~A controller in the \hrt game \emph{Praxis Fighter X}, verbatim: it locates the ship and nearby bullets by color, dodges while firing, and hands control back when its loop ends. It played 103 steps until the win in \cref{fig:teaser}d.}
\label{lst:controller}
\vspace{-1em}
\end{figure*}

\section{Related Work}
\label{sec:related}

\cref{tab:related} compares \ours with representative agents and harnesses, and \cref{sec:supp-related} gives an extended discussion.

\paragraph{Harnesses for visual and game agents.}
Most agent harnesses follow ReAct~\cite{yao2023react}, taking one action per reasoning step.
Later work changes what the model sees or which tools it may call~\cite{yang2024sweagent,lin2023swiftsage}.
On ARC-AGI-3~\cite{arcagi3}, \vista~\cite{han2026vista} lifts the same models to perfect scores through visual memory and re-inspection of past frames.
OpenVLHarness~\cite{zhou2026openvlharness} adds visual tools and multimodal memory for static visual reasoning.
Embodied agent frameworks and harnesses let general vision-language models act in simulated environments and control robots~\cite{yang2025embodiedbench,chen2026showharness,li2026motormind}.
Computer-use agents~\cite{qin2025uitars,xie2024osworld} and game agents~\cite{tan2025cradle,wang2025gametars,tan2025lumine,jiang2026game2player} also act on screenshots.
Most game agents, however, pause the game while the model generates, as do many game benchmarks~\cite{ying2026aigamestore} and the Lite setting of VideoGameBench~\cite{zhang2025videogamebench}.
These harnesses improve what the model sees or remembers, but assume that the world waits for it.

\paragraph{Evaluation in real time.}
Latency is a major reason why vision-language models fail at real-time games~\cite{zhang2025videogamebench}, and recent benchmarks keep the environment running during inference~\cite{wen2026realtime,wang2026roar,tang2026realtimebench,kang2025winfast}.
Closest to ours, The World Won't Wait~\cite{pan2026worldwontwait} compares paused and live clocks and finds that a strong but slow model wins only when the game is paused.
Its agents, however, read structured text states.
Its TORSI method still calls a model for every decision, with a slow teacher guiding a fast student.
\ours acts on pixels and instead changes what one model call can do.

\paragraph{Acting between model calls.}
Temporally extended actions date back to options~\cite{sutton1999options}.
Language models also write programs that serve as policies or skills~\cite{liang2023codeaspolicies,wang2024voyager,wang2024codeact,karten2026continualharness}, but these programs mostly run on structured state in environments that wait.
AutoHarness~\cite{lou2026autoharness} refines a code policy with environment feedback and resets offline, per game and for turn-based text games.
Real-time agents pair a slow model with a fast layer: a reactive thread~\cite{wen2026realtime}, concurrent inference coroutines~\cite{yakushev2026async}, action chunks~\cite{black2025rtc}, options held until a timer or repetition check fires~\cite{kaiya2023lyfe}, a scripted executor over fixed macro actions~\cite{liu2024hla}, rule-based reflexes with latency charged to the simulation clock~\cite{zheng2025reflex}, or reflexes and policy trees compiled in advance~\cite{chen2026typego,dong2026late}.
Asynchronous function calling lets a returning call interrupt the model~\cite{gim2024asynclm}.
The most similar mechanism is DPT-Agent~\cite{zhang2025dptagent}, in which the model writes condition rules at run time that steer a hand-built state machine over the symbolic state of one game.
In these systems, the fast path is mostly another model inference, code fixed in advance, or rules for a game-specific executor.
AsyncLLM can rewrite its coroutines, but only between evaluation rounds.
\ours instead \emph{delegates} acting to controllers that the model writes from pixels during the episode, including their perception, and \emph{interrupts} them through wake events.

\section{AutoPILOT}
\label{sec:method}

\ours is developed on \vista~\cite{han2026vista}, which is built on the Codex harness, with additional tools for running macros and controllers in the background and wake events that return control to the model.
As in \vista, the model sees rendered frames together with the game clock and its remaining budgets, but never the game state or the DOM, from which only restart and decision-screen events are derived (\cref{sec:supp-signals}).
\ours keeps \vista's \tool{act} tool, read-only tools (\tool{inspect}, \tool{read\_pixels}, \tool{history}) and note files, so the two harnesses expose the same tools apart from delegation.
All games share the same harness instructions and only the game description, task and controls differ. \cref{sec:supp-prompt} shows the full prompt, which adds budget rules, play tips and delegation rules to \vista's instructions.
\ours also reports the remaining budget after each tool result and notifies the model of restarts.

\subsection{Acting in real time}
\label{sec:setting}

At time $t$ the agent can read a frame $o(t)$ and send an \emph{action}: one keyboard or mouse event or an explicit wait, which counts as one \emph{environment step}.
The model is invoked in discrete \emph{model calls}: call $k$ reads the frame at time $t_k$ and takes $\ell_k$ seconds, a median of 6.1\,s for \gpt at low effort.
Existing visual harnesses turn each model call into one action, computed from $o(t_k)$ but executed at $t_k+\ell_k$.
In real time this causes two problems: every action is $\ell_k$ seconds stale, so a projectile that was safe in $o(t_k)$ may have arrived by then, and the agent acts at most once per model call.
Pausing the clock removes both but changes the task.

\ours instead lets a single model call act over many steps, by starting a macro or a controller that keeps acting without further model calls.
The model thus chooses how many steps each model call covers, and its actions no longer wait for the next call.

\subsection{Three execution modes}
\label{sec:modes}

As shown in \cref{fig:teaser}b, each model call may invoke one of three tools that act on the game.
They differ in how long they act without the model and whether they read the screen.

\noindent\textbf{Primitive action (\tool{act}).}
The model sends one action. \tool{act} also stops any running macro or controller.

\noindent\textbf{Open-loop macro (\tool{run\_macro}).}
The model writes a sequence of actions, \eg a menu route learned from earlier interaction, and the worker sends them without looking.

\noindent\textbf{Closed-loop controller (\tool{python}).}
The model writes a short Python function \texttt{run(api)} that the worker executes for up to 120\,s in a sandbox with one CPU, 1\,GB of memory and no network access.
\cref{sec:supp-harness} gives the full setup.
It reads frames with \texttt{api.observe()}, decides with ordinary code, and acts with \texttt{api.act()}.
A controller re-reads the frame and acts with no model call.
Each step takes about 0.4\,s.
The model decides in its code how long each action lasts.
This averages about 0.2\,s.
Taking the next screenshot and reading the state take the other 0.2\,s.
It can also ask the model for help through a \emph{handoff}, one of the wake events described in \cref{sec:handoff}.
\cref{lst:controller} shows a macro and a controller from two recorded episodes.

\subsection{Handing control back}
\label{sec:handoff}

For delegation to be safe, control must return to the model once the running program no longer fits the situation.
While a macro or controller runs, the model's tool call blocks: the model waits instead of producing tokens, and the game is played by the worker.
The call returns with a fresh frame on one of six \emph{wake events}:
\begin{itemize}[leftmargin=*,itemsep=1pt,topsep=2pt,parsep=0pt]
\item \textbf{From the program.} \emph{Handoff}: the controller requests one with a reason, \eg on reaching a goal or seeing something it was not written for. It may keep playing while the model reads the reason. \emph{Return}: \texttt{run} ends at the stopping condition the model wrote or with an error. \emph{Macro done}: all macro actions are sent.
\item \textbf{From the harness.} \emph{Decision screen}: the game waits for a choice, such as in a shop. \emph{Scene change}: most of the screen differs from a second earlier and then holds still. The harness then stops the program, since its plan may no longer fit. \emph{Timeout}: the program has run for 120\,s.
\end{itemize}
A game restart after a loss affects the two modes differently.
It stops a running macro, which then returns and wakes the model.
It does not wake the model while a controller runs, since the controller observes an incremented \texttt{reset\_epoch} and can keep playing.
Restarts do not pause the game: the next attempt starts at once, whether the model is thinking or a program is running. A running controller keeps playing it, and the model learns of the failure at its next wake event, where it can reflect and write a revised controller.
After waking, the model can keep, replace or stop (with \tool{act}) the running controller.
Thinking and acting overlap only after a handoff, while the controller keeps playing and the model reads the reason.
This overlap is small in practice (2.3--3.5\% of inference time, \cref{sec:supp-ablations}).

\section{AFK-Bench}
\label{sec:bench}
\begin{wrapfigure}[26]{R}{0.60\linewidth}
  \vspace{-1.4em}
  \centering
  \includegraphics[width=\linewidth]{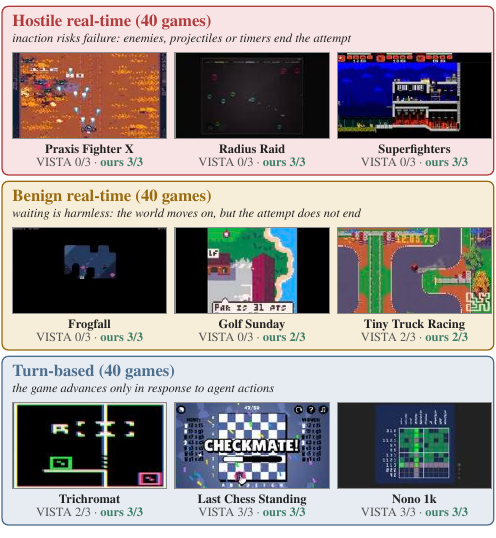}
  \vspace{-1.6em}
  \caption{\textbf{The three environment types of \bench.} Each row shows three of the 40 games of one type. Below each game are the real-time wins of \vista and \ours out of three episodes with \gpt. \cref{tab:games} lists all 120 games.}
  \label{fig:bench}
  \vspace{-1em}
\end{wrapfigure}

\bench contains three environment types that vary the cost of inaction, and two clock protocols that play the same games in real time and paused, to test whether an agent acts well in real time.

\paragraph{Games and tasks.}
\bench consists of 120 browser games, each with one task and an official success check, played in a real browser.
The games span shooters, platformers, arcade and action games, puzzles and board-like games.
\cref{sec:supp-bench} describes how we collected them from public game sites and validated each with 13 scripted checks.
Success and progress are computed from the game state by the benchmark's evaluator.
This state is never exposed to the agent, which sees only rendered frames plus restart and decision-screen events (\cref{sec:supp-signals}).

\paragraph{Environment types.}
We divide the games into three types of 40 by what happens when the agent does nothing.
To assign each game, a scripted probe stops all actions for 60\,s and checks whether the attempt fails, the world keeps changing, or nothing happens.
\cref{fig:bench} shows three example games of each type, together with how often \vista and \ours win them in real time.
\par\noindent\textbullet~\emph{\Hrt environments}, where inaction risks failure: an idle agent loses the attempt, health or a life. In the shooter \emph{Praxis Fighter X}, a ship that does not dodge is destroyed within seconds.
\par\noindent\textbullet~\emph{\Brt environments}, where the world keeps moving but inaction causes no direct harm: an idle agent loses time, not the attempt, as in \emph{Tiny Truck Racing}.
\par\noindent\textbullet~\emph{\Tb environments}, where the game changes only when the agent acts and no countdown runs, as in the box-pushing puzzle \emph{Trichromat}.
\par
The types thus separate two costs of slow thinking: in \hrt environments it causes failure, while in the other two it only wastes time and budget.
\cref{sec:supp-bench} details the probe, including a second probe for games whose danger appears only after some play.

\paragraph{Clock protocols.}
Every game can be played under two clock protocols.
In \emph{real time}, the game clock never stops: the game keeps running during each model call, as for a human player who looks away from the keyboard.
Under the \emph{paused clock}, the harness freezes the game from the start of each model call until it ends, together with any running macro or controller.
Game time therefore passes only while no model call is in progress.
Every game becomes turn-based for the agent, with no time pressure while the model thinks.
Under both protocols, the games' own pause and restart controls are blocked, so the agent cannot stop the clock itself.
The drop from the paused clock to real time measures how much an agent loses to its own latency.

\paragraph{Budgets.}
An episode ends with a win, the game's own end, or when one of its budgets is used up: \textbf{100 model calls}, \textbf{2000\,s of game time} or \textbf{2000 environment steps}.
A model call is one invocation of the model, which issues at most one tool call (a reply without one also counts).
An environment step is one action sent to the game, whether by \tool{act}, a macro or a controller.
A single-step harness thus spends one model call per step and sends at most 100 steps, so only the call and time budgets bind it, whereas one \ours model call can launch a macro or controller that sends many steps, and the step budget caps these.
In real time, thinking time counts as game time, while loading and automatic restarts do not.
After a loss, the game restarts within the same budgets, so the agent can try again.

\paragraph{Metrics.}
Our main metric is the success rate (SR): the fraction of episodes in which the game's official success check fires before any budget is used up.
Because an episode can span several attempts, SR measures whether an agent wins within its budgets, possibly after earlier losses.
\cref{sec:why} also reports success on the first attempt.
The mean progress (PG) gives a finer signal.
It averages over episodes the highest progress reached in any attempt, which the evaluator normalizes to $[0,1]$ for each game.
We also report the number of model calls, tokens and API cost per episode.

\section{Experiments}
\label{sec:exp}

We aim to answer four questions: how much a single-step harness loses when the world keeps running and whether \ours recovers it (\cref{sec:main}), where the gains come from (\cref{sec:why}), how they scale with budget, effort and speed (\cref{sec:robust}), and whether they generalize (\cref{sec:general}).

\begin{table*}[t]
  \centering
  \small
  \setlength{\tabcolsep}{6pt}
  \begin{tabular}{llcccc}
    \toprule
    Harness & Model & \Hrt & \Brt & \Tb & Overall \\
    \midrule
    \rowcolor{gray!13}\multicolumn{6}{c}{\textbf{Paused Clock}} \\
    Codex only & \gpt & 52.5 \prog{737} & 45.0 \prog{693} & 57.5 \prog{780} & 51.7 \prog{737} \\
    Claude Code only & \opus & 47.5 \prog{735} & 40.0 \prog{594} & 40.0 \prog{664} & 42.5 \prog{665} \\
    \cmidrule{1-6}
     & \gpt & 56.7 \prog{783} & 55.8 \prog{746} & 55.8 \prog{796} & 56.1 \prog{775} \\
     & \opus & 37.5 \prog{629} & 40.0 \prog{644} & 40.0 \prog{670} & 39.2 \prog{648} \\
     & DeepSeek-V4.1-Flash & 25.0 \prog{537} & 22.5 \prog{468} & 25.0 \prog{496} & 24.2 \prog{500} \\
     & Qwen4-Max & 32.5 \prog{552} & 27.5 \prog{513} & 15.0 \prog{445} & 25.0 \prog{503} \\
     & GLM-5.3-Flash & 2.5 \prog{240} & 0.0 \prog{233} & 2.5 \prog{130} & 1.7 \prog{201} \\
    \multirow{-6}{*}{\vista} & Hy4-flash & 12.5 \prog{298} & 2.5 \prog{272} & 2.5 \prog{131} & 5.8 \prog{233} \\
    \cmidrule{1-6}
    \oursrow & \gpt & 73.3 \prog{857} & 63.3 \prog{795} & 82.5 \prog{915} & \best{73.1} \prog{856} \\
    \oursrow & \opus & 65.0 \prog{795} & 40.0 \prog{663} & 77.5 \prog{892} & \best{60.8} \prog{784} \\
    \oursrow & DeepSeek-V4.1-Flash & 50.0 \prog{663} & 32.5 \prog{531} & 50.0 \prog{692} & \best{44.2} \prog{629} \\
    \oursrow & Qwen4-Max & 50.0 \prog{653} & 30.0 \prog{515} & 45.0 \prog{718} & \best{41.7} \prog{628} \\
    \oursrow & GLM-5.3-Flash & 35.0 \prog{588} & 20.0 \prog{453} & 25.0 \prog{513} & \best{26.7} \prog{518} \\
    \oursrow \multirow{-6}{*}{\ours} & Hy4-flash & 45.0 \prog{658} & 30.0 \prog{538} & 40.0 \prog{688} & \best{38.3} \prog{628} \\
    \midrule
    \rowcolor{gray!13}\multicolumn{6}{c}{\textbf{Real-time Clock}} \\
     & No-op & 0.0 \prog{079} & 0.0 \prog{001} & 0.0 \prog{002} & 0.0 \prog{027} \\
    \multicolumn{1}{c}{\multirow{-2}{*}{---}} & Random & 10.0 \prog{360} & 2.5 \prog{214} & 5.0 \prog{199} & 5.8 \prog{258} \\
    \cmidrule{1-6}
    Codex only & \gpt & 17.5 \prog{469} & 36.7 \prog{594} & 62.5 \prog{811} & 38.9 \prog{625} \\
    Claude Code only & \opus & 20.0 \prog{486} & 25.0 \prog{544} & 52.5 \prog{729} & 32.5 \prog{586} \\
    \cmidrule{1-6}
     & \gpt & 16.7 \prog{442} & 37.5 \prog{595} & 60.8 \prog{801} & 38.3 \prog{613} \\
     & \opus & 17.5 \prog{453} & 27.5 \prog{502} & 45.0 \prog{733} & 30.0 \prog{563} \\
     & DeepSeek-V4.1-Flash & 10.0 \prog{313} & 25.0 \prog{506} & 25.0 \prog{576} & 20.0 \prog{465} \\
     & Qwen4-Max & 7.5 \prog{298} & 10.0 \prog{377} & 25.0 \prog{528} & 14.2 \prog{401} \\
     & GLM-5.3-Flash & 5.0 \prog{226} & 2.5 \prog{188} & 2.5 \prog{072} & 3.3 \prog{162} \\
    \multirow{-6}{*}{\vista} & Hy4-flash & 7.5 \prog{220} & 0.0 \prog{186} & 2.5 \prog{157} & 3.3 \prog{188} \\
    \cmidrule{1-6}
    \oursrow & \gpt & 62.5 \prog{784} & 60.0 \prog{777} & 83.3 \prog{912} & \best{68.6} \prog{824} \\
    \oursrow & \opus & 47.5 \prog{671} & 42.5 \prog{657} & 70.0 \prog{882} & \best{53.3} \prog{737} \\
    \oursrow & DeepSeek-V4.1-Flash & 40.0 \prog{642} & 30.0 \prog{539} & 47.5 \prog{667} & \best{39.2} \prog{616} \\
    \oursrow & Qwen4-Max & 25.0 \prog{540} & 25.0 \prog{484} & 32.5 \prog{606} & \best{27.5} \prog{543} \\
    \oursrow & GLM-5.3-Flash & 12.5 \prog{421} & 10.0 \prog{428} & 32.5 \prog{516} & \best{18.3} \prog{455} \\
    \oursrow \multirow{-6}{*}{\ours} & Hy4-flash & 30.0 \prog{532} & 25.0 \prog{477} & 42.5 \prog{716} & \best{32.5} \prog{575} \\
    \bottomrule
  \end{tabular}
  \vspace{-0.5em}
  \caption{\textbf{Main results on \bench.} Success rate (\%) with mean progress in small type. Bold: best overall success rate for each model and clock. DeepSeek-V4.1-Flash, GLM-5.3-Flash and Hy4-flash are open-source, and Qwen4-Max is closed. All models use low reasoning effort. Hy4-flash has a single thinking mode. \cref{tab:supp-ci} gives 95\% intervals and statistics over the three seeds.}
  \label{tab:main}
  \vspace{-1em}
\end{table*}

\subsection{Setup}
\label{sec:setup}

\paragraph{Models.}
We compare all harnesses with two closed models.
We access \gpt through the Codex app server~\cite{openai2026gpt61} and \opus through the Claude Code CLI~\cite{anthropic2026opus55}.
\vista and \ours also run with three open-source models, DeepSeek-V4.1-Flash~\cite{deepseek2026v41flash}, GLM-5.3-Flash~\cite{zai2026glm53flash} and Hy4-flash~\cite{tencent2026hy4flash}.
They also run with one more closed model, Qwen4-Max~\cite{qwen2026qwen4max}.
We access these four models through OpenAI-compatible APIs.
\cref{sec:supp-models} adds paired tests for these four models and GameWorld results for Qwen4-Max and Hy4-flash.
It also runs \ours with GPT-6 Astra~\cite{openai2026gpt6astra} and Claude Fable 5.2~\cite{anthropic2026fable52}.
All main runs use low reasoning effort. Hy4-flash has a single thinking mode.
\cref{sec:robust} compares low and high effort for \gpt and \opus.

\paragraph{Baselines.}
We compare against three visual agent harnesses.
\emph{\vista}~\cite{han2026vista} is the previous state-of-the-art visual harness on ARC-AGI-3~\cite{arcagi3} and GameWorld.
We run its official implementation through the same Codex app server, with its own prompt and tools: lossless visual memory, visual inspection, notes, and one action per \tool{act} call.
\emph{Codex} and \emph{Claude Code} are the plain harnesses of the two models, with only \tool{act} and \tool{screenshot} as game tools, and represent a standard single-step computer-use loop.
We tell every harness that the game runs in real time.
\cref{sec:supp-baselines} gives the exact instructions.
Methods that read text states~\cite{wen2026realtime,pan2026worldwontwait,lou2026autoharness} or require training~\cite{tan2025lumine,wang2025gametars} do not apply to this screen-only, training-free setting.
We additionally include two simple policies without a model to validate the benchmark: a \emph{no-op agent} that sends no step, and a \emph{random agent} that sends random valid steps as fast as possible.

\paragraph{Protocol.}
Unless noted otherwise, results are in real time on all 120 tasks. For \gpt, we run three episodes per task with different random seeds (360 in total).
To limit cost, we run one episode per task for the other models.

\subsection{Main results}
\label{sec:main}

\paragraph{\ours outperforms existing harnesses.}
As \cref{tab:main} shows, \ours nearly doubles the real-time success rate of existing harnesses: 68.6\% against 38.3\% for \vista and 38.9\% for Codex with \gpt, and 53.3\% against 30.0\% and 32.5\% for \vista and Claude Code with \opus, while costing \$0.89 per episode against \$1.50 for \vista with \gpt (\cref{tab:cost}).
The margin is largest in \hrt environments (62.5\% \vs 16.7\% with \gpt) and remains large in the other two types.
The no-op and random agents win at most 5.8\%.
Of the 360 paired \gpt episodes, 111 are won only by \ours and 2 only by \vista.

\paragraph{Open-source models and Qwen4-Max.}
The three open-source models and the closed Qwen4-Max show the same pattern at lower success rates.
With each of them, \ours outperforms \vista under both clocks, by 13.3 (Qwen4-Max) to 29.2 points (Hy4-flash) in real time, and every difference is significant (\cref{sec:supp-models}).
\ours with DeepSeek-V4.1-Flash (39.2\%) surpasses \vista with \gpt (38.3\%), and \ours with Hy4-flash (32.5\%) surpasses \vista with \opus (30.0\%), so \textbf{a better harness can make up for a weaker model}.
The reason \vista falls behind is not the same for the open-source models as for \gpt and \opus.
With \gpt and \opus, it loses much of its success to the clock (17.8 and 9.2 points).
With the three open-source models, it is already weak when the game waits (1.7--24.2\%) and loses at most 4.2 points in real time.
With Qwen4-Max, it reaches 25.0\% when the game waits and loses 10.8 points in real time.
With GLM-5.3-Flash and Hy4-flash, \vista wins almost no tasks under either clock, while \ours wins 18.3--38.3\%.
In relative terms, the gain is larger with each of these four models: \ours reaches 1.9--9.8$\times$ the real-time success rate of \vista, against 1.8$\times$ with \gpt and \opus.
When a model plays poorly one action at a time, letting it write a whole plan helps all the more.

\paragraph{Memory and perception tools bring no overall gain.}
In real time, \vista performs no better than the plain Codex and Claude Code harnesses (38.3\% \vs 38.9\% with \gpt, 30.0\% \vs 32.5\% with \opus), despite its lossless visual memory and inspection tools.
Each \vista action is computed from a frame about one model call old, so a better view of it cannot help.
Even with the clock paused, these tools add 4.4 points over Codex (56.1\% \vs 51.7\%) and none over Claude Code (39.2\% \vs 42.5\%).

\paragraph{\ours is less sensitive to the game clock.}
Moving from the paused clock to real time lowers the success rate of \vista and Codex by 17.8 and 12.8 points with \gpt, but that of \ours by 4.5.
The drops are closer with \opus (9.2 and 10.0 for \vista and Claude Code, 7.5 for \ours), while on GameWorld \vista loses 41.2 points and \ours only 2.9 (\cref{sec:general}).

\begin{figure*}[t]
  \centering
  \includegraphics[width=0.93\textwidth]{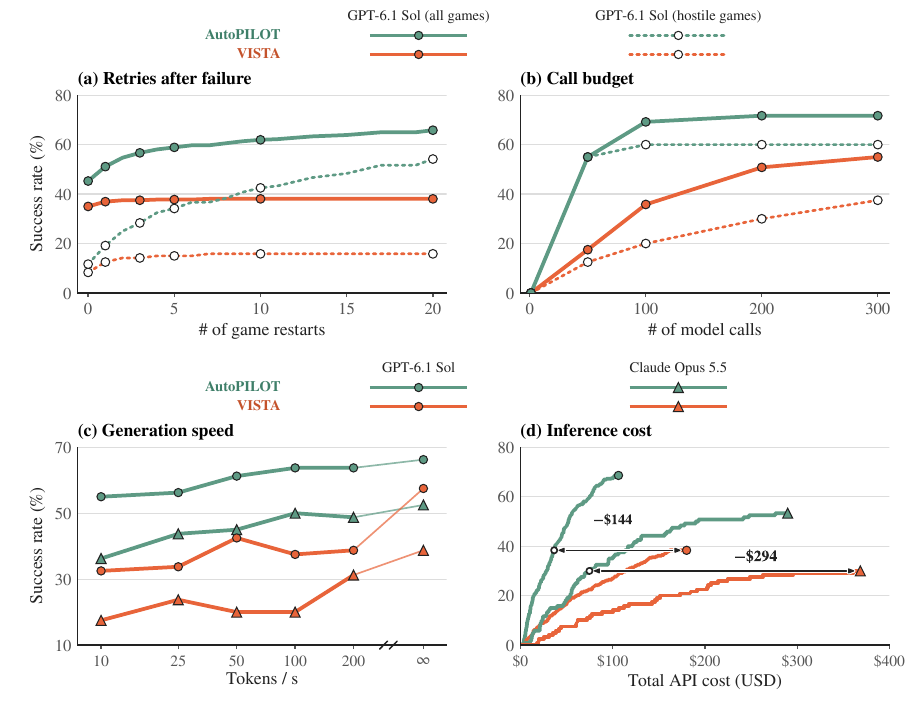}
  \vspace{-0.8em}
  \caption{\textbf{Recovery, model-call budget, generation speed and cost.}
  \textbf{(a)}~Success within $k$ restarts and \textbf{(b)}~within the first $b$ model calls (300-call budget), on all 120 games (solid) and the 40 \hrt games (dotted), with \gpt.
  \textbf{(c)}~Success of \ours and \vista on the 80 \hrt and \brt tasks against simulated generation speed (in tokens/s, where $\infty$ means a paused clock), with \gpt and \opus. \textbf{(d)}~Success against the total list-price API cost over all 120 tasks as a per-episode spending cap is raised. Arrows run from where \ours first matches \vista's final success (white ring) to \vista's total cost.}
  \label{fig:curves}
\end{figure*}

\subsection{Why \ours wins}
\label{sec:why}

We analyze the gap in three steps: gains beyond latency, the responsible component, and recovery from failures.
\paragraph{Gains beyond latency.}
Even with the clock paused, where no action is stale, \ours beats \vista by 17.0 points with \gpt and 21.7 with \opus, so \emph{latency explains only part of the gap}.
The rest comes from \textbf{executing a whole plan per model call instead of a single action}, which makes far better use of the model-call budget.
\vista spends one model call on each action, so long action sequences use up its budget: 208 of its 222 real-time failures end by exhausting its 100 calls, after only 78 steps per episode.
\ours instead sends 681 steps from 43 model calls, about 16 per call, because one macro or controller carries out many steps.
\emph{The model also chooses the mode that fits the game} (\cref{tab:tools}): controllers send 98\% of the steps in \hrt and 89\% in \brt environments, whereas in \tb environments, which allow planning ahead, macros account for 36\% of calls and 26\% of steps.

\paragraph{Which component is responsible.}
\cref{tab:cost} removes the two delegation tools, \tool{python} and \tool{run\_macro}, one at a time (seeds in \cref{tab:supp-seeds}).
Without controllers, which reduces \ours to batching actions into macros, success in \hrt environments drops from 62.5\% to 50.8\%, with fewer steps and more model calls.
Without macros, overall success drops slightly, mainly in \tb environments, where macros are used most (83.3\% to 78.3\%, \cref{tab:cost-full}).
With either tool alone, \ours still clearly outperforms \vista, so \textbf{the two modes can largely stand in for each other}, while combining them further improves overall success with the fewest tokens (2.8M).
Controllers that the model writes from its own view but that cannot read the screen while running win slightly fewer tasks than closed-loop ones (78 \vs 83, \cref{sec:supp-ablations}).
Removing both drops \ours to 45.8\% overall and 22.5\% in \hrt environments.
This variant keeps \vista's tools and adds only our prompt, budget line and restart notices. It still outperforms \vista on the same seed, but \emph{these additions matter far less than the two tools}, which account for 28 of the 38 extra wins overall and 17 of 19 in \hrt environments.

\begin{table*}[t]
  \centering
  \small
  \setlength{\tabcolsep}{3pt}
  {%
  \begin{tabular}{@{}lcccccccc@{}}
    \toprule
    \multirow{2}{*}{Env.} & \multicolumn{4}{c}{\% of tool calls} & \multicolumn{3}{c}{\% of env steps} & \multirow{2}{*}{\makecell{Steps\\per game}} \\
    \cmidrule(lr){2-5}\cmidrule(lr){6-8}
     & \tool{act} & macro & ctrl. & other & \tool{act} & macro & ctrl. & \\
    \midrule
    Hostile & 37 & 10 & 46 & 7 & 1 & 1 & 98 & 1043 \\
    Benign & 46 & 25 & 25 & 4 & 4 & 7 & 89 & 693 \\
    Turn-based & 43 & 36 & 17 & 4 & 6 & 26 & 68 & 307 \\
    \midrule
    All & 43 & 25 & 27 & 5 & 3 & 7 & 90 & 681 \\
    \bottomrule
  \end{tabular}}
  \vspace{-0.6em}
  \captionof{table}{\textbf{Tool calls and steps of \ours} (\gpt, real time). Share of tool calls and environment steps by tool (each sums to 100\%), and steps per game. Macros and controllers (ctrl.) send many steps per call. ``Other'' covers note and inspection tools.}
    \label{tab:tools}
  \vspace{0.8em}

  \small
  \setlength{\tabcolsep}{3pt}
  {%
  \begin{tabular}{@{}lccccc@{}}
    \toprule
    & \multicolumn{2}{c}{Success (\%)} & & & \\
    \cmidrule(lr){2-3}
    Harness & Hostile & Overall & Calls & Tokens & \$/ep. \\
    \midrule
    Codex & 17.5 & 38.9 & 81 & 5.1M & 1.35 \\
    \vista & 16.7 & 38.3 & 80 & 5.3M & 1.50 \\
    \midrule
    \oursrow \ours & \best{62.5} & \best{68.6} & \best{43} & \best{2.8M} & \best{0.89} \\
    \quad No controller & 50.8 & 64.7 & 58 & 4.6M & 1.31 \\
    \quad No macro & \best{62.5} & 67.5 & 46 & 3.1M & 0.91 \\
    \quad No controller \& macro & 22.5 & 45.8 & 81 & 5.5M & 1.45 \\
    \bottomrule
  \end{tabular}}
  \vspace{-0.6em}
  \captionof{table}{\textbf{Ablations and cost} (\gpt, real time). Success rate (\%) and model calls, tokens and cost per episode.}
    \label{tab:cost}
\end{table*}

\paragraph{Recovering from failures.}
Restarts are where \ours gains most (\cref{fig:curves}a).
Without any restart, \ours wins 11.7\% of \hrt episodes and \vista 8.3\%.
With up to 20 restarts, \ours climbs to 54.2\% and is still rising, whereas \vista stalls at 15.8\% after about seven restarts.
\vista must \emph{still play every retry one action per model call}, so each attempt meets the same latency, whereas \ours can play it at game speed in a macro or controller.
An attempt costs \ours 2 model calls against 8 for \vista with \gpt (1 \vs 12 with \opus, \cref{fig:lives}).
After a failure, \ours therefore still has budget to retry.
It wins 58\% of paired episodes against 6\% for \vista, which had used up its calls in 110 of 123 (\cref{sec:supp-restarts}).

\subsection{Scaling with budget, effort and speed}
\label{sec:robust}

\begin{table}[t]
  \centering
  \setlength{\tabcolsep}{7pt}
  \newcommand{\blim}[1]{#1}
  \begin{tabular}{llcc}
    \toprule
    Budget & \blim{model calls\,/\,steps\,/\,time} & Success & Cost / ep. \\
    \midrule
    \rowcolor{gray!13}\multicolumn{4}{c}{\textbf{\ours}} \\
    Standard & \blim{100\,/\,2000\,/\,2000\,s} & 69.2 & \$0.83 \\
    3$\times$ model calls & \blim{300\,/\,2000\,/\,2000\,s} & 71.7 (+2.5) & \$1.18 \\
    2$\times$ steps & \blim{100\,/\,4000\,/\,2000\,s} & 70.8 (+1.7) & \$0.90 \\
    2$\times$ game time & \blim{100\,/\,2000\,/\,4000\,s} & 70.8 (+1.7) & \$0.94 \\
    All relaxed & \blim{300\,/\,4000\,/\,4000\,s} & \best{78.3} (+9.2) & \$1.92 \\
    \rowcolor{gray!13}\multicolumn{4}{c}{\textbf{\vista}} \\
    Standard & \blim{100\,/\,--\,/\,2000\,s} & 37.5 & \$1.48 \\
    3$\times$ model calls & \blim{300\,/\,--\,/\,2000\,s} & 55.0 (+17.5) & \$5.28 \\
    All relaxed & \blim{300\,/\,--\,/\,4000\,s} & 51.7 (+14.2) & \$6.53 \\
    \bottomrule
  \end{tabular}
  \vspace{-0.6em}
  \caption{\textbf{Scaling the budget.} Success rate (\%) and API cost per episode with \gpt in real time on 120 episodes. Parentheses give the change from the standard budget. \vista sends one step per model call, so the step budget does not apply~(--).}
    \label{tab:budgets-main}
  \vspace{-1em}
\end{table}
\paragraph{Scaling the budget.}
Since \ours sends far more steps per episode, we test whether \vista simply runs out of budget too early (\cref{tab:budgets-main,tab:supp-budgets}).
Relaxing the budgets does help \vista: with three times the model calls, it rises from 37.5\% to 55.0\%.
Yet a large gap remains, as \ours reaches 69.2\% at the standard budget and 78.3\% with all budgets relaxed, and in \hrt environments \vista stays at or below 37.5\% under every budget.
\vista's gain is costly: \$5.28 per episode, 6$\times$ that of \ours, which reaches 55\% within 50 calls (\cref{fig:curves}b).
Capping the spending per episode (\cref{fig:curves}d), \ours matches \vista's final success rate at 0.2$\times$ its total cost with both models (\$37 \vs \$180 with \gpt, \$75 \vs \$369 with \opus).

\begin{figure}[t]
  \centering
  \includegraphics[width=0.68\linewidth]{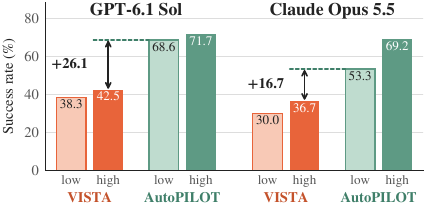}
  \caption{\textbf{Low \vs high reasoning effort} (\ours and \vista, real time). Overall success rate. Arrows mark the gap from \vista at high effort to \ours at low effort, in points.}
  \label{fig:effort}
\end{figure}
\paragraph{Scaling reasoning effort.}
Prior work finds that slower, reasoning-heavy agents degrade in real time, as the world moves on while the model thinks~\cite{wen2026realtime,pan2026worldwontwait}.
In our runs, high effort hurts neither harness, but only \ours turns it into a large gain (\cref{fig:effort}).
With \opus, \ours rises from 53.3\% to 69.2\% and \vista from 30.0\% to 36.7\%.
With \gpt, both gain a few points (68.6\% to 71.7\% and 38.3\% to 42.5\%).
At either effort, \vista stays far below \ours at low effort, by 26.1 and 16.7 points at high effort.
For \ours, a longer call yields a better program that then acts for many steps.

\paragraph{Scaling generation speed.}
We simulate faster or slower models with a decoding speed of $v$ tokens/s: the clock stops during generation and then advances by $N/v$\,s for $N$ output tokens.
On the 80 \hrt and \brt tasks (\cref{fig:curves}c), \ours with \gpt at low effort rises from 55.0\% at 10 tokens/s to 63.8\% at 200 tokens/s, close to the paused clock.
\vista, by contrast, reaches 38.8\% even at 200 tokens/s, against 57.5\% when paused.
\ours stays strong even at low speed because \textbf{speed only delays when a working controller starts}: once written, a controller plays at full speed however slowly the model thinks.
Slow generation hurts most when outputs are long: \opus at high effort, which writes about twice as many tokens per call as \gpt, gains the most from speed (31.3\% to 62.5\%, \cref{tab:speed}).

\Needspace{9\baselineskip}
\subsection{Generalization to GameWorld}
\label{sec:general}
\begin{wraptable}{r}{0.56\linewidth}
  \vspace{-1.2em}
  \centering
  \small
  \setlength{\tabcolsep}{4.5pt}
  \newcommand{\ap}{\cellcolor{apbg}}
  \begin{tabular}{llcc}
    \toprule
    Model & Harness & Paused & Real time \\
    \midrule
     & \vista & \best{84.1} \prog{917} & 42.9 \prog{653} \\
    \multirow{-2}{*}{\gpt} & \ap\ours & \ap 82.9 \prog{911} & \ap\best{80.0} \prog{900} \\
    \cmidrule{1-4}
     & \vista & 67.1 \prog{830} & 41.2 \prog{623} \\
    \multirow{-2}{*}{\opus} & \ap\ours & \ap\best{72.4} \prog{855} & \ap\best{67.6} \prog{845} \\
    \bottomrule
  \end{tabular}
  \vspace{-0.6em}
  \caption{\textbf{Results on GameWorld.} Success rate (\%) on its 170 tasks, with mean progress in small type. Bold marks the best per model and clock.}
    \label{tab:gameworld}
  \vspace{-1em}
\end{wraptable}

We also evaluate both harnesses on GameWorld~\cite{ouyang2026gameworld}, whose 170 tasks in 34 browser games do not overlap with \bench, using the same budgets (\cref{tab:gameworld}).
With the clock paused, the two harnesses are close (84.1\% \vs 82.9\% with \gpt), as GameWorld tasks are short enough that \vista rarely runs out of calls (3 of its 27 failures).
In real time, however, \vista loses 41.2 points with \gpt and 25.9 with \opus, while \ours loses only 2.9 and 4.7.
The \textbf{robustness of \ours to the game clock carries over to a different set of games}.

\section{Conclusion}
\label{sec:conclusion}

We presented \ours, a visual agent harness in which the model delegates acting to macros and controllers that it writes at run time.
Because each model call can launch a whole plan rather than a single action, \ours substantially outperforms existing harnesses in real time with closed and open-source models and on two benchmarks, without any training.
With \gpt, it also costs less.
Our results suggest that how model outputs become actions over time deserves as much attention in harness design as what the model sees and remembers, for game agents and agents operating live interfaces or robots.

\bibliographystyle{plainnat}
\bibliography{main}

\clearpage\setcounter{mainpages}{\value{page}}
\clearpage
\setcounter{page}{1}
\maketitlesupplementary
\appendix

\section{Extended related work}
\label{sec:supp-related}

\paragraph{Agent harnesses.}
A harness is the code around a model that turns the environment into the model's observations and the model's outputs into actions.
ReAct~\cite{yao2023react} interleaves reasoning with one action per turn.
SWE-agent~\cite{yang2024sweagent} shows that the interface given to the model matters as much as the model.
CodeAct~\cite{wang2024codeact} lets a single action be a program that calls many tools.
AgentSPEX~\cite{wang2026agentspex} specifies workflow branches and loops, and Lean4Agent~\cite{wang2026lean4agent} checks the consistency of these workflows in Lean.
\vista~\cite{han2026vista} perceives the game as images, keeps every frame in a lossless visual memory and lets the model re-inspect frames as it reasons, raising the ARC-AGI-3 score of the same model from 40.68 to 100.
Our episodes restart after a loss, so the model can reflect on failed attempts as in Reflexion~\cite{shinn2023reflexion}.

\paragraph{Computer-use and game agents.}
Vision-based computer-use agents ground actions in screenshots~\cite{cheng2024seeclick,qin2025uitars,lu2026qwencua,xie2024osworld,yang2026openwebrl}.
Game agents range from reinforcement-learning policies trained per game~\cite{mnih2015dqn} to foundation-model agents that play many games from pixels or structured state~\cite{wang2024voyager,wang2025jarvis1,simateam2024sima,simateam2025sima2,tan2025cradle,wang2025gametars}.
Lumine~\cite{tan2025lumine} trains a vision-language model to play 3D open-world games in real time.
It reads frames at 5\,Hz, emits actions at 30\,Hz and reasons only when needed.
It reaches real-time play through training, whereas \ours uses a general model without training.

\paragraph{Real-time evaluation.}
VideoGameBench~\cite{zhang2025videogamebench} reports latency as a major failure cause for vision-language models.
Real-Time Reasoning Gym~\cite{wen2026realtime} and ROAR~\cite{wang2026roar} advance their environments during inference.
Realtime-Bench~\cite{tang2026realtimebench} evaluates 30 web and 20 game tasks from screenshots under wall, paused and token clocks with a one-action-per-model-call harness.
Win Fast or Lose Slow~\cite{kang2025winfast} adapts model size and quantization to the time pressure.
The World Won't Wait~\cite{pan2026worldwontwait} evaluates 180 tasks from six game types.
It uses structured text observations and game-specific action interfaces.
Several of these plan ahead, such as button plans in Mario and flap schedules in Bird.
While a model call is pending, the previous control stays active.

\paragraph{Acting while thinking.}
AgileThinker~\cite{wen2026realtime} runs a reactive and a planning thread in parallel, and SwiftSage~\cite{lin2023swiftsage} pairs a small fine-tuned model with a large one, in the spirit of dual-process theories~\cite{kahneman2011thinking}.
AsyncLLM~\cite{yakushev2026async} is a general framework for asynchronous inference rather than a game agent, and it is evaluated mainly on streaming video understanding.
In two ViZDoom scenarios, a background thinker sets a course of action that a fast coroutine turns into per-frame actions from the two latest frames.
An appendix experiment lets the agent rewrite its own coroutines between evaluation rounds.
Real-time chunking~\cite{black2025rtc} executes one action chunk of a robot policy while the next is generated.
Dong \etal~\cite{dong2026late} show that GUI agents often choose correctly but act too late.
They pre-compile guarded policy trees sized to the decoding latency, which raises success within the decision window from 0.50 to 0.79.
Their branches are enumerated in advance for discrete GUI states, whereas \ours's model writes closed-loop programs for continuous games at run time.
TypeGo~\cite{chen2026typego} runs planning loops at several timescales and has an LLM compile reflexes offline into Python condition and action functions that run at 100\,Hz.
\ours synthesizes its controllers at run time from the current screen.

\FloatBarrier
\section{Harness details}
\label{sec:supp-harness}

\paragraph{Tools.}
\cref{tab:supp-tools} lists the tools available to the model in each harness.
\ours keeps \vista's \tool{act} and read-only tools and adds \tool{run\_macro} and \tool{python}, which together give the three execution modes of \cref{sec:modes}.

\begin{table}[!htbp]
  \centering
  \footnotesize
  \setlength{\tabcolsep}{3pt}
  \begin{tabular}{@{}lccc@{}}
    \toprule
    Tool & Codex & \vista & \ours \\
    \midrule
    \tool{act} (one action per call) & \checkmark & \checkmark & \checkmark \\
    \tool{run\_macro} (open-loop) & & & \checkmark \\
    \tool{python} (closed-loop controller) & & & \checkmark \\
    \tool{inspect}, \tool{read\_pixels}, \tool{history} & & \checkmark & \checkmark \\
    notes (\texttt{GUIDE.md}, \texttt{WORKING.md}) & & \checkmark & \checkmark \\
    \bottomrule
  \end{tabular}
  \caption{Game tools per harness in the \gpt runs. Every harness sees a screenshot after each action, and \ours also on every wake. The Codex harness also has \tool{screenshot} and keeps Codex's code execution without shell, file or network access.}
  \label{tab:supp-tools}
\end{table}

\paragraph{Actions.}
Every action, whether sent by \tool{act}, a macro or a controller, uses one schema.
Its types are \texttt{wait}, \texttt{press\_key}, \texttt{press\_keys}, \texttt{type}, \texttt{click}, \texttt{click\_hold}, \texttt{mouse\_move}, \texttt{drag} and \texttt{scroll}.
The action \texttt{press\_key} holds one key, and \texttt{press\_keys} holds two keys at once but rejects opposing directions.
The action \texttt{type} enters up to five characters.
Each action is one environment step.
Coordinates refer to the $1280\times720$ screenshot, durations are at most 2\,s, and the default hold or wait is 0.4\,s.
Controllers usually set shorter durations.
Only the keys of the game's declared controls are allowed.

\paragraph{Macros.}
\tool{run\_macro} takes a list of at most 200 actions and sends them in order without reading the screen.
With the default \texttt{on\_reset: stop}, a game restart stops the macro.
With \texttt{continue}, the list restarts on the new attempt.

\paragraph{Controller API.}
A controller is a function \texttt{run(api)} that a background worker executes for at most 120\,s.
It runs in a Docker sandbox with one CPU, 1\,GB of memory, 60\,s of CPU time, no network and no access to the game or host files.
Python, NumPy and a \texttt{run\_macro} helper are available.
\texttt{api.observe()} returns the latest frame as a $720\times1280\times3$ RGB array together with the public state: the game clock, the remaining budgets and \texttt{reset\_epoch}.
\texttt{api.act(action)} sends one action in the format above, which counts as one step toward the step budget.
\texttt{api.request\_handoff(reason)} wakes the model while the controller keeps running.
Returning from \texttt{run} ends the controller and also wakes the model.
\texttt{api.notify\_reset(reason)} records a game restart, signalled by an incremented \texttt{reset\_epoch}, without waking the model.
A new controller replaces the running one, and \tool{act} stops it.
There is no fixed control rate.
A controller observes and acts as fast as its own code and action durations allow.
\cref{sec:supp-ablations} reports the measured step time.

\paragraph{Waiting and waking.}
After a \tool{run\_macro} or \tool{python} call, the harness holds the tool result while the macro or controller runs, so the model produces no tokens in the meantime.
It polls every 0.25\,s for one of the wake events of \cref{sec:handoff}.
The result is released with a fresh frame on a handoff, when the program ends, stops or times out, or when the episode ends.

\paragraph{Screen watch.}
Each macro and controller has a screen watch, which raises the scene-change wake event and which the model can tune or turn off per call.
A \emph{scene change} fires when at least half of the screen differs from the frame about one second earlier and the new screen then holds for three frames, so a flash or continuous motion does not trigger it.
An optional \emph{static} check fires when the screen stays unchanged for 10\,s while actions are sent.
On a trigger the watch stops the program by default.
It can instead only wake the model.

\paragraph{Signals derived from the game state.}
\label{sec:supp-signals}
No harness reads the game state, but two kinds of events derived from it reach the agent.
\emph{Automatic restarts.} After a loss, \ours's controllers see an incremented \texttt{reset\_epoch}, and the model receives a one-line notice that the attempt failed and the game restarted.
Codex receives only this notice, and \vista receives the last frame before the restart.
\emph{Decision screens.} Four of the 120 tasks, all \hrt, declare game states in which the game waits for a choice.
In real time, \ours and Codex poll the game state every 0.25\,s.
When such a screen appears, the running controller is stopped and the model receives a neutral notice.
\vista receives no such notice.
No harness receives it with the clock paused.
Removing the four tasks changes the real-time gap between \ours and \vista from +30.3 to +30.2 points with \gpt and from +23.3 to +21.6 with \opus.

\FloatBarrier
\section{Prompt}
\label{sec:supp-prompt}
Each prompt starts with the game's description, task and controls, which are shared by all harnesses.
\vista then receives its official instructions, with one added sentence in real time: \emph{``Real time continues while you think and while tools run; the game is not paused between your actions.''}
\ours receives the same text followed by the rules block below, shown in its real-time version for the \gpt runs.
The prompts are reproduced verbatim and call environment steps ``game inputs''.
Under the paused clock, the first paragraph instead states that the game is paused while the model thinks, and the last rule that a running controller is paused while the model thinks.

\begin{lstlisting}[basicstyle=\ttfamily\scriptsize,breaklines=true,columns=fullflexible]
# Real-time play

The game keeps running while you think, call tools, and run Python.

## Budget and ending
- You have 2000 seconds of game time. Loading and automatic restarts are not counted.
- You can make at most 100 tool calls; a response without a tool call also counts as one. Actions sent by your running controller do not count toward this limit. All game inputs together are limited to 2000.
- A response without a tool call is submitted as a final answer and the episode continues. Only the game decides success.

## Choosing a tool
- Use `act` for single inputs: starting, quick reactions and recovery.
- Calling `act` stops a running controller.
- Use `run_macro` for a short sequence whose effect you have already seen.
- Use `python` for a controller that keeps playing from the screen while you think. Prefer `run_macro` when it is enough.
- Test an uncertain control briefly before you rely on it for longer.
- Decide what visible change counts as progress. Motion on screen and passing time are not progress.
- Let a controller keep running while you think only if it is already working. Otherwise have it hand control back and stop.
\end{lstlisting}

The Codex harness receives no instructions of ours beyond the game text and the following rules:
\begin{lstlisting}[basicstyle=\ttfamily\scriptsize,breaklines=true,columns=fullflexible]
- The game runs in real time. It does not pause while you think or use tools.
- You have 2000 seconds of game time. Loading and automatic restarts are not counted.
- You can send at most 100 responses. Each `screenshot` or `act` call counts as one response, and so does a reply with no tool call.
- You can send at most 2000 game inputs. Each `act` is one input, including waits.
- A reply with no tool call is submitted as a final answer and the episode continues; it also uses one game input. The episode ends only on success, game end or budget end.
\end{lstlisting}
Later runs of \ours add three rules and end every tool result with a budget line such as \texttt{Budget left: 1999 of 2000 game inputs, 99 of 100 responses, 1989 of 2000 s game time.}
\begin{lstlisting}[basicstyle=\ttfamily\scriptsize,breaklines=true,columns=fullflexible]
- Each `act`, each `run_macro` action and each `api.act` call uses one game input, including waits. A response without a tool call can also use one game input. Other tools and `api.observe()` use none.
- If any budget remains, keep playing: try a different approach instead of giving a final answer.
- Combine several waits, clicks or key presses into one `run_macro` call: each tool call uses one of your 100 responses.
\end{lstlisting}
Later runs of Codex and \vista add the keep-playing rule and a similar budget line.

\FloatBarrier
\section{Baselines}
\label{sec:supp-baselines}
\cref{sec:supp-prompt} gives the real-time instructions of \vista and the Codex harness.
\vista runs its official GameWorld profile and tools through the Codex app server of Codex CLI 0.159.2, like \ours.
Its budget is 100 tool calls plus 2000\,s of game time.
Every tool call counts, including \tool{inspect}, \tool{read\_pixels}, \tool{history} and the note tools.
The Codex harness runs Codex's own base instructions, agent loop and context compaction in a single thread per episode.
Web search, browser and computer use, skills, plugins and sub-agents are disabled.
Its code execution has no shell, file system or network access.
The Claude Code harness runs Claude Code's default system prompt and agent loop with \tool{act} and \tool{screenshot} as its only tools.
All built-in tools are disabled, and context-compaction turns do not count as model calls.
The no-op agent sends no step from the start of an episode to its end, with the same automatic restarts and the same evaluator.
Each episode runs for 600\,s of game time.
It wins no task.
Its mean progress is .079 in \hrt environments and at most .002 in the other two, consistent with the type assignment of \cref{sec:bench}.
The random agent sends a uniformly random valid step as fast as the harness allows, at a median of 2.9 steps per second.
It runs one episode per task, within 2000 steps and 2000\,s.
It wins 5.8\% of tasks, mostly in \hrt environments, where it wins 10.0\%.
Its mean progress is .258, against .027 for the no-op agent, so progress is best read against this floor.

\FloatBarrier
\section{Benchmark details}
\label{sec:supp-bench}

\paragraph{Games.}
The 120 games run offline in a pinned Chromium image.
Of these, 37 are built with HTML5 canvas, 21 with PICO-8, 22 with Unity WebGL, 17 with Flash through the Ruffle emulator, 13 with Godot, 9 with other WebAssembly engines and 1 with Phaser.
Each game exposes its state to the evaluator through a small per-game adapter, which the agent cannot access.
None of the 34 games of GameWorld is among them.
We collected and wrapped all 120 games ourselves from freely playable web games.
The manifest records the source of 112 of them and pins every game file by its hash.
Of these, 94 are on itch.io, 10 on Armor Games, 3 on Kongregate, 3 on Newgrounds, 1 on Not Doppler and 1 on GitHub.
\cref{tab:game-sources} lists each game's author, source and licence.
Eight games carry a licence stated by their author.
The other 112 games state no licence, so we treat their terms as unknown.

\paragraph{Selection.}
From a pool of 568 collected games, each game was assigned an environment type from its description and idle probes.
A task was admitted only after a scripted acceptance test that uses no model and runs with the same code and browser image as the reported runs.
Its 13 checks test whether a scripted policy can win in real time.
They also require that pausing through either harness stops the game clock and that resuming restores normal speed.
Score, prompt and screen must agree, the game must show no undeclared freeze, and failures and restarts must be detected.
Memory must stay within its limit, and 60\,s of idling must cause no penalty in \tb games but must cause one in \hrt games.
The game's own pause must be blocked, and phases the agent cannot control must be recognized.
The budget must be feasible, the prompt must name only allowed keys, and the game state must be sampled at a regular rate.
The 120 tasks were frozen before any run reported here.
For games whose danger appears only after some play, a second probe first plays a scripted policy for up to 120\,s and then idles for 60\,s.

\paragraph{Episode rules.}
All runs block the games' own pause and restart controls.
From the moment a game reports a failure until it is playable again, all actions are discarded but still counted as steps, including those of a running controller.
This time does not count as game time, and neither does loading or an automatic restart.
Phases the agent cannot control, such as automatic cutscenes, do not count as game time either.
A wall-clock safety cap of 4000\,s ends any paused episode.
Every real-time episode is audited against the game's own clock for pauses.
An episode is scored by the benchmark's evaluator only.
A run that fails for infrastructure reasons is retried automatically and not scored.
A used-up budget is a failure.
A model response that takes longer than the per-call deadline of 300\,s, or a request that the model provider rejects as too large, counts as a failure of the episode.
\cref{tab:games} lists all 120 games with their environment type and per-game results.

\begin{table*}[p]
  \centering
  \scriptsize
  \setlength{\tabcolsep}{2.6pt}
  \renewcommand{\arraystretch}{0.98}
    \begin{minipage}[t]{0.485\textwidth}
  \centering
  \begin{tabular}[t]{@{}>{\raggedright\arraybackslash}p{0.62\linewidth}cc@{}}
    \toprule
    \multicolumn{3}{@{}l}{{\bfseries \Hrt (40)}} \\
    Game & \vista & \ours \\
    \midrule
    A Dance of Fire and Ice & 0/3 & {\bfseries 3/3} \\
    Backstreet Warriors & 0/3 & 0/3 \\
    Bad Monday Simulator & 0/3 & {\bfseries 2/3} \\
    Cave Buster & 0/3 & {\bfseries 1/3} \\
    Dead Estate CLASSIC & 0/3 & {\bfseries 3/3} \\
    Deadbeats Encore & 0/3 & 0/3 \\
    driftr & 1/3 & 1/3 \\
    Eat not, Want not & 0/3 & {\bfseries 3/3} \\
    Enigmata 2: Genu's Revenge & 3/3 & 3/3 \\
    Epic Coaster & 0/3 & 0/3 \\
    Fair Squares & 1/3 & {\bfseries 3/3} \\
    Flooded Caves & 3/3 & 3/3 \\
    Galactic Wars & 2/3 & {\bfseries 3/3} \\
    Generic Fighter Maybe & 0/3 & {\bfseries 1/3} \\
    Grave & 0/3 & {\bfseries 3/3} \\
    Gun Mayhem & 0/3 & 0/3 \\
    HIT8OX & 2/3 & {\bfseries 3/3} \\
    Just One Boss & 0/3 & {\bfseries 1/3} \\
    Lumberjack Run & 3/3 & 3/3 \\
    MOBS, INC & 0/3 & {\bfseries 3/3} \\
    Monolith & 0/3 & 0/3 \\
    My Rusty Submarine & 0/3 & 0/3 \\
    O ($\odot$) & 0/3 & {\bfseries 3/3} \\
    Packed to the Gills & 0/3 & 0/3 \\
    PortaBoy & 1/3 & {\bfseries 3/3} \\
    PRAXIS FIGHTER X & 0/3 & {\bfseries 3/3} \\
    Project: Intellisphere & 0/3 & {\bfseries 1/3} \\
    Radius Raid & 0/3 & {\bfseries 3/3} \\
    RetroFab & 1/3 & {\bfseries 3/3} \\
    return to sender & 0/3 & 0/3 \\
    SCARLET SHIFT & 0/3 & 0/3 \\
    Seraph's Last Stand & 0/3 & {\bfseries 3/3} \\
    Sharpest Spear & 0/3 & {\bfseries 3/3} \\
    Sir Truck & 0/3 & {\bfseries 1/3} \\
    Smash 'n Dab & 0/3 & {\bfseries 1/3} \\
    Strike Force Heroes & 3/3 & 3/3 \\
    Superfighters & 0/3 & {\bfseries 3/3} \\
    Tenjutsu 48h & 0/3 & {\bfseries 2/3} \\
    UZG & 0/3 & {\bfseries 1/3} \\
    XENOPUNCH & 0/3 & {\bfseries 3/3} \\
    \midrule
    Total & 20/120 & 75/120 \\
    \bottomrule
  \end{tabular}
  \end{minipage}\hfill
  \begin{minipage}[t]{0.485\textwidth}
  \centering
  \begin{tabular}[t]{@{}>{\raggedright\arraybackslash}p{0.62\linewidth}cc@{}}
    \toprule
    \multicolumn{3}{@{}l}{{\bfseries \Brt (40)}} \\
    Game & \vista & \ours \\
    \midrule
    Achievement Unlocked & 3/3 & 3/3 \\
    APOLLO & 0/3 & 0/3 \\
    CELESTE Classic & 0/3 & 0/3 \\
    Chronotron & 3/3 & 3/3 \\
    Crabjuice & 0/3 & {\bfseries 2/3} \\
    Dank Tomb & 0/3 & 0/3 \\
    Debbie's Diner Derby & 0/3 & {\bfseries 3/3} \\
    Dusk Child & 0/3 & 0/3 \\
    Egg & 0/3 & 0/3 \\
    Eggdog Extended & 0/3 & 0/3 \\
    Exit Path & 0/3 & {\bfseries 1/3} \\
    Frogfall & 0/3 & {\bfseries 3/3} \\
    Give Up & 0/3 & 0/3 \\
    Golf Sunday & 0/3 & {\bfseries 2/3} \\
    Interdimensional Spike & 3/3 & 3/3 \\
    King of Drift & 3/3 & 3/3 \\
    Lo-Fi Room & 0/3 & 0/3 \\
    Looper & 3/3 & 3/3 \\
    Moss Moss & 3/3 & 3/3 \\
    Mr.\ Platformer & 0/3 & 0/3 \\
    Offroaders 2 & 3/3 & 3/3 \\
    \"Ooo (demo) & 3/3 & 3/3 \\
    pico sonic & 0/3 & {\bfseries 2/3} \\
    PolyTrack & 0/3 & 0/3 \\
    PUP & 2/3 & {\bfseries 3/3} \\
    Ray Ardent Science Ninja & 3/3 & 3/3 \\
    Reversal Rooms & 2/3 & {\bfseries 3/3} \\
    SELF & 0/3 & {\bfseries 3/3} \\
    SHIFT 2 & 3/3 & 3/3 \\
    SHRUBNAUT & 0/3 & 0/3 \\
    Sky Run & 2/3 & {\bfseries 3/3} \\
    Smol Ame & 0/3 & 0/3 \\
    speedloop & 0/3 & {\bfseries 3/3} \\
    Sulka & 1/3 & {\bfseries 3/3} \\
    TEMPRES & 0/3 & {\bfseries 3/3} \\
    The Sun And Moon & 3/3 & 3/3 \\
    This Is The Only Level & 3/3 & 3/3 \\
    Tiny Truck Racing & 2/3 & 2/3 \\
    UFO Swamp Odyssey & 0/3 & 0/3 \\
    William and Sly & 0/3 & 0/3 \\
    \midrule
    Total & 45/120 & 72/120 \\
    \bottomrule
  \end{tabular}
  \end{minipage}
  \caption{\textbf{The 120 games of \bench}, by environment type, with real-time wins out of three episodes with \gpt for \vista and \ours (bold: higher). Titles follow the games' source pages. \cref{tab:game-sources} lists sources and licences.}
  \label{tab:games}
\end{table*}
\begin{table*}[p]
  \ContinuedFloat
  \centering
  \scriptsize
  \setlength{\tabcolsep}{2.6pt}
  \renewcommand{\arraystretch}{0.98}
  \begin{minipage}[t]{0.485\textwidth}
  \centering
  \begin{tabular}[t]{@{}>{\raggedright\arraybackslash}p{0.62\linewidth}cc@{}}
    \toprule
    \multicolumn{3}{@{}l}{{\bfseries \Tb (40)}} \\
    Game & \vista & \ours \\
    \midrule
    AGLOBO & 0/3 & {\bfseries 1/3} \\
    awake & 3/3 & 3/3 \\
    Blackout & 0/3 & 0/3 \\
    Bloody Fun Day & 1/3 & {\bfseries 2/3} \\
    Bob Velseb the Cannibal the Dating Simulator & 3/3 & 3/3 \\
    Cavern Sweeper & 3/3 & 3/3 \\
    Chessformer & 0/3 & {\bfseries 1/3} \\
    Claustrowordia & 3/3 & 3/3 \\
    Combo Pool & 0/3 & {\bfseries 3/3} \\
    Coming Out Simulator 2014 & 3/3 & 3/3 \\
    Covemount & 0/3 & 0/3 \\
    Curious Fishing & 2/3 & {\bfseries 3/3} \\
    Dungeon Raid & 0/3 & {\bfseries 3/3} \\
    FARA & {\bfseries 3/3} & 2/3 \\
    Fossil Quest & 3/3 & 3/3 \\
    Gnomitaire & 3/3 & 3/3 \\
    Karawan & 3/3 & 3/3 \\
    Katti & 3/3 & 3/3 \\
    Kitsu & {\bfseries 2/3} & 1/3 \\
    Last Chess Standing & 3/3 & 3/3 \\
    Lowland Keepers & 0/3 & {\bfseries 3/3} \\
    Mai-Chan's Sweet Buns & 0/3 & {\bfseries 3/3} \\
    Mole Mole & 0/3 & {\bfseries 2/3} \\
    Nip For Speed & 3/3 & 3/3 \\
    Nono 1k & 3/3 & 3/3 \\
    PancitoMerge & 3/3 & 3/3 \\
    Patient Rogue & 3/3 & 3/3 \\
    Peggy's Post & 0/3 & 0/3 \\
    Pegote & 3/3 & 3/3 \\
    Roguewood & 1/3 & {\bfseries 3/3} \\
    Roulette Knight & 0/3 & {\bfseries 2/3} \\
    Sebastian's Quest & 0/3 & {\bfseries 2/3} \\
    Six-Sided Streets & 2/3 & {\bfseries 3/3} \\
    Snug & 2/3 & {\bfseries 3/3} \\
    Streamline & 3/3 & 3/3 \\
    The Pigeon Post Principle & 3/3 & 3/3 \\
    Tiny Tower Defense & 2/3 & {\bfseries 3/3} \\
    Trichromat & 2/3 & {\bfseries 3/3} \\
    Vanguard Monsters & 3/3 & 3/3 \\
    Woodworm Classic & 2/3 & {\bfseries 3/3} \\
    \midrule
    Total & 73/120 & 100/120 \\
    \bottomrule
  \end{tabular}
  \end{minipage}
  \caption[]{\textbf{The 120 games of \bench} (continued).}
\end{table*}
\begin{table*}[p]
  \centering
  \scriptsize
  \setlength{\tabcolsep}{2.2pt}
  \renewcommand{\arraystretch}{0.93}
  \begin{minipage}[t]{0.325\textwidth}
  \centering
  \begin{tabular}[t]{@{}>{\raggedright\arraybackslash}p{0.37\linewidth}>{\raggedright\arraybackslash\hspace{0pt}}p{0.26\linewidth}cc@{}}
    \toprule
    Game & Author & Src. & Lic. \\
    \midrule
    Achievement Unlocked & John & A & -- \\
    A Dance of Fire and Ice & fizzd et al. & I & -- \\
    AGLOBO & Daniel Moreno & I & -- \\
    APOLLO & Daniel Moreno et al. & I & -- \\
    awake & droqen & I & -- \\
    Backstreet Warriors & SebaGamesDev & I & -- \\
    Bad Monday Simulator & Lumpy et al. & I & -- \\
    Blackout & FRESH & I & -- \\
    Bloody Fun Day & urbansquall & K & -- \\
    Bob Velseb the Cannibal the Dating Simulator & numijulie & I & -- \\
    Cave Buster & Jon Topielski et al. & I & -- \\
    Cavern Sweeper & Hempuli & I & -- \\
    CELESTE Classic & Maddy Makes Games et al. & I & -- \\
    Chessformer & rob1221 & I & -- \\
    Chronotron & -- & N & -- \\
    Claustrowordia & anttihaavikko & I & -- \\
    Combo Pool & NuSan & I & NC-SA \\
    Coming Out Simulator 2014 & Nicky Case & I & PD \\
    Covemount & Hempuli & I & -- \\
    Crabjuice & dev\_dwarf et al. & I & -- \\
    Curious Fishing & -- & -- & -- \\
    Dank Tomb & Jakub Wasilewski & I & -- \\
    Deadbeats Encore & fie & I & -- \\
    Dead Estate CLASSIC & Milkbar Lads & I & -- \\
    Debbie's Diner Derby & -- & -- & -- \\
    driftr & kindanice & I & -- \\
    Dungeon Raid & 4Cats & I & -- \\
    Dusk Child & Sophie Houlden & I & -- \\
    Eat not, Want not & Sitrex et al. & I & -- \\
    Egg & Terry Cavanagh & I & -- \\
    Eggdog Extended & zamsire et al. & I & -- \\
    Enigmata 2: Genu's Revenge & kidgamez & K & -- \\
    Epic Coaster & knigg & K & -- \\
    Exit Path & John & A & -- \\
    Fair Squares & kindanice & I & -- \\
    FARA & BrianIsCreative & I & -- \\
    Flooded Caves & NuSan & I & NC-SA \\
    Fossil Quest & Unsigned Double & I & -- \\
    Frogfall & Kultisti & I & -- \\
    Galactic Wars & Volcano Bytes & I & NC-ND \\
    \bottomrule
  \end{tabular}
  \end{minipage}\hfill
  \begin{minipage}[t]{0.325\textwidth}
  \centering
  \begin{tabular}[t]{@{}>{\raggedright\arraybackslash}p{0.37\linewidth}>{\raggedright\arraybackslash\hspace{0pt}}p{0.26\linewidth}cc@{}}
    \toprule
    Game & Author & Src. & Lic. \\
    \midrule
    Generic Fighter Maybe & Astrobard Games et al. & I & -- \\
    Give Up & John & A & -- \\
    Gnomitaire & Arnold Rauers & I & -- \\
    Golf Sunday & Johan Peitz & I & -- \\
    Grave & Heartbeast & I & -- \\
    Gun Mayhem & KevinGu & A & -- \\
    HIT8OX & vladcomarlau & I & -- \\
    Interdimensional Spike & yaotu & I & -- \\
    Just One Boss & ayla\textasciitilde nonsense & I & NC-SA \\
    Karawan & -- & -- & -- \\
    Katti & Kultisti & I & -- \\
    King of Drift & -- & A & -- \\
    Kitsu & F\'ayer et al. & I & -- \\
    Last Chess Standing & FeatureKreep & I & -- \\
    Lo-Fi Room & Bearmask & I & -- \\
    Looper & FeatureKreep & I & -- \\
    Lowland Keepers & Twotinydice & I & -- \\
    Lumberjack Run & terrasima & I & MIT \\
    Mai-Chan's Sweet Buns & Krystman & I & -- \\
    MOBS, INC & overboy & I & -- \\
    Mole Mole & Adam Saltsman & I & -- \\
    Monolith & samloeschen & I & -- \\
    Moss Moss & Noel Cody & I & -- \\
    Mr.\ Platformer & -- & -- & -- \\
    My Rusty Submarine & S{\o}rb et al. & I & -- \\
    Nip For Speed & Knackelibang et al. & I & -- \\
    Nono 1k & NuSan & I & -- \\
    O ($\odot$) & sirmilkman & I & -- \\
    Offroaders 2 & ArmorGames & A & -- \\
    \"Ooo (demo) & NamaTakahashi & I & -- \\
    Packed to the Gills & moonlight\-jammers et al. & I & -- \\
    PancitoMerge & F\'ayer & I & -- \\
    Patient Rogue & watabou & I & -- \\
    Peggy's Post & digitarium & I & -- \\
    Pegote & jvolonte et al. & I & -- \\
    pico sonic & komehara & I & MIT/NC \\
    PolyTrack & Kodub & I & -- \\
    PortaBoy & -- & -- & -- \\
    PRAXIS FIGHTER X & SPARSE//\allowbreak GameDev & I & -- \\
    Project: Intellisphere & stylemistake & I & -- \\
    \bottomrule
  \end{tabular}
  \end{minipage}\hfill
  \begin{minipage}[t]{0.325\textwidth}
  \centering
  \begin{tabular}[t]{@{}>{\raggedright\arraybackslash}p{0.37\linewidth}>{\raggedright\arraybackslash\hspace{0pt}}p{0.26\linewidth}cc@{}}
    \toprule
    Game & Author & Src. & Lic. \\
    \midrule
    PUP & mmatt\_ugh & I & -- \\
    Radius Raid & Jack Rugile & G & MIT \\
    Ray Ardent Science Ninja & shaneneville & A & -- \\
    RetroFab & Itizso & I & -- \\
    return to sender & kindanice & I & -- \\
    Reversal Rooms & TaelsDaFoox & I & -- \\
    Roguewood & watabou & I & -- \\
    Roulette Knight & Four Quarters team & I & -- \\
    SCARLET SHIFT & sirmilkman et al. & I & -- \\
    Sebastian's Quest & Adam Saltsman & I & -- \\
    SELF & dev\_dwarf et al. & I & -- \\
    Seraph's Last Stand & -- & -- & -- \\
    Sharpest Spear & NPJarcade & I & -- \\
    SHIFT 2 & Tony & A & -- \\
    SHRUBNAUT & gleeson & I & -- \\
    Sir Truck & -- & -- & -- \\
    Six-Sided Streets & Chris Klimowski & I & -- \\
    Sky Run & bikas & A & -- \\
    Smash 'n Dab & Will Herring & I & -- \\
    Smol Ame & MooCow Games & I & -- \\
    Snug & Cookiecrayon & I & -- \\
    speedloop & kindanice et al. & I & -- \\
    Streamline & Francois van Niekerk & I & -- \\
    Strike Force Heroes & Sky9 Games & D & -- \\
    Sulka & Kultisti & I & -- \\
    Superfighters & -- & N & -- \\
    TEMPRES & tak & I & -- \\
    Tenjutsu 48h & Deepnight Games & I & -- \\
    The Pigeon Post Principle & cassowary & I & -- \\
    The Sun And Moon & Daniel Linssen & I & -- \\
    This Is The Only Level & John & A & -- \\
    Tiny Tower Defense & BuilderBot & I & -- \\
    Tiny Truck Racing & -- & -- & -- \\
    Trichromat & spratt & I & -- \\
    UFO Swamp Odyssey & Paranoid Cactus & I & -- \\
    UZG & nex & I & -- \\
    Vanguard Monsters & Jay & I & -- \\
    William and Sly & -- & N & -- \\
    Woodworm Classic & spratt & I & -- \\
    XENOPUNCH & overboy & I & -- \\
    \bottomrule
  \end{tabular}
  \end{minipage}
  \caption{\textbf{Sources and licences of the 120 games.} Author: the first author or uploader named on the source page. Src.: I itch.io, A Armor Games, K Kongregate, N Newgrounds, D Not Doppler, G GitHub, -- no source page recorded. Lic.: the licence the author states on the source page, in the source repository or in a licence file shipped with the game. MIT is the MIT licence (code), NC, NC-SA and NC-ND are CC BY-NC 4.0, CC BY-NC-SA 4.0 and CC BY-NC-ND 4.0 (assets or cartridge), MIT/NC is MIT code with CC BY-NC 4.0 assets, PD is a public-domain dedication, and -- means that no licence is stated (unknown).}
  \label{tab:game-sources}
\end{table*}

\FloatBarrier
\section{Additional results}
\label{sec:supp-results}

\FloatBarrier
\subsection{Main results and their reliability}
\label{sec:supp-main}

\paragraph{Runs.}
The configurations with \gpt in \cref{tab:supp-seeds} and \vista with \gpt under the paused clock consist of three runs over all 120 tasks.
All other configurations are run once per task.

\begin{table}[!htbp]
  \centering
  \footnotesize
  \setlength{\tabcolsep}{4pt}
  \begin{tabular}{@{}lcccc@{}}
    \toprule
    Harness (real time, \gpt) & run 1 & run 2 & run 3 & mean \\
    \midrule
    Codex & 39.2 & 37.5 & 40.0 & 38.9 \\
    \vista & 37.5 & 39.2 & 38.3 & 38.3 \\
    \ours & 69.2 & 68.3 & 68.3 & 68.6 \\
    \quad No controller & 66.7 & 65.0 & 62.5 & 64.7 \\
    \quad No macro & 65.0 & 70.0 & 67.5 & 67.5 \\
    \ours, high effort & 71.7 & 72.5 & 70.8 & 71.7 \\
    \ours, direct API & 68.3 & 65.8 & 69.2 & 67.8 \\
    \midrule
    \ours, paused clock & 71.7 & 75.8 & 71.7 & 73.1 \\
    \bottomrule
  \end{tabular}
  \caption{Success rate (\%) in each of the three runs, with random seeds 42, 43 and 44. Each run covers 120 tasks.}
  \label{tab:supp-seeds}
\end{table}

\paragraph{Intervals and paired tests.}
\cref{tab:supp-ci} gives a 95\% bootstrap interval over tasks for the overall success rate of each harness with \gpt and \opus, and exact McNemar tests on paired episodes of the same task for the comparisons in \cref{sec:exp}.

\begin{table*}[!tp]
  \centering
  \footnotesize
  \setlength{\tabcolsep}{3pt}
  \begin{tabular}{@{}llccc@{}}
    \toprule
    Model, clock & Harness & SR (\%) & 95\% interval & Run std \\
    \midrule
    \gpt, paused & \vista & 56.1 & [48.3, 63.9] & 1.0 \\
     & \ours & 73.1 & [65.8, 80.0] & 2.4 \\
    \gpt, real time & Codex & 38.9 & [30.8, 47.2] & 1.3 \\
     & \vista & 38.3 & [30.3, 46.4] & 0.8 \\
     & \ours & 68.6 & [61.1, 76.1] & 0.5 \\
    \opus, paused & \vista & 39.2 & [30.8, 48.3] & -- \\
     & \ours & 60.8 & [51.7, 69.2] & -- \\
    \opus, real time & \vista & 30.0 & [21.7, 38.3] & -- \\
     & \ours & 53.3 & [44.2, 62.5] & -- \\
    \bottomrule
  \end{tabular}

  \vspace{0.6em}
  \begin{tabular}{@{}lrrrl@{}}
    \toprule
    Comparison (A \vs B) & Pairs & A only & B only & $p$ \\
    \midrule
    \multicolumn{5}{@{}l}{\emph{\gpt, \bench}} \\
    \ours \vs \vista, real time & 360 & 111 & 2 & $10^{-30}$ \\
    Codex \vs \vista, real time & 360 & 18 & 16 & 0.86 \\
    \ours \vs \vista, paused & 360 & 71 & 10 & $2\times10^{-12}$ \\
    \vista paused \vs real time & 360 & 82 & 18 & $6\times10^{-11}$ \\
    \ours paused \vs real time & 360 & 34 & 18 & 0.036 \\
    \ours \vs No controller & 360 & 28 & 14 & 0.044 \\
    \ours \vs No macro & 360 & 23 & 19 & 0.64 \\
    High \vs low effort & 360 & 27 & 16 & 0.13 \\
    Direct API \vs Codex app server & 360 & 20 & 23 & 0.76 \\
    No controller \& macro \vs \vista & 120 & 12 & 2 & 0.013 \\
    \quad \hrt only & 40 & 3 & 1 & 0.62 \\
    \ours \vs No controller \& macro & 120 & 30 & 2 & $2.5\times10^{-7}$ \\
    \quad \hrt only & 40 & 18 & 1 & $8\times10^{-5}$ \\
    Relaxed \vs standard budget & 120 & 13 & 2 & 0.007 \\
    4000\,s \vs 2000\,s & 120 & 7 & 5 & 0.77 \\
    4000 \vs 2000 steps & 120 & 9 & 7 & 0.80 \\
    \multicolumn{5}{@{}l}{\emph{\opus, \bench}} \\
    \ours \vs \vista, real time & 120 & 30 & 2 & $2.5\times10^{-7}$ \\
    \ours \vs \vista, paused & 120 & 30 & 4 & $6\times10^{-6}$ \\
    High \vs low effort, real time & 120 & 24 & 5 & $5\times10^{-4}$ \\
    \multicolumn{5}{@{}l}{\emph{\gpt, GameWorld}} \\
    \ours \vs \vista, real time & 170 & 63 & 0 & $2\times10^{-19}$ \\
    \ours paused \vs real time & 170 & 9 & 4 & 0.27 \\
    \ours \vs \vista, paused & 170 & 7 & 9 & 0.80 \\
    \vista paused \vs real time & 170 & 72 & 2 & $3\times10^{-19}$ \\
    \bottomrule
  \end{tabular}
  \caption{\textbf{Intervals and paired tests.} Top: success rate with a 95\% bootstrap interval over tasks and the std over the three runs. The bootstrap uses 4{,}000 resamples and averages the episodes of a task. A dash marks a single run. Bottom: pairs won by only one side, and the two-sided exact McNemar $p$.}
  \label{tab:supp-ci}
\end{table*}

\paragraph{By game genre.}
\cref{tab:supp-genre} breaks the environment types of \cref{tab:main} down into game genres.

\begin{table}[!htbp]
  \centering
  \footnotesize
  \setlength{\tabcolsep}{2.3pt}
  \begin{tabular}{lccccc}
    \toprule
    & & \multicolumn{2}{c}{\gpt} & \multicolumn{2}{c}{\opus} \\
    \cmidrule(lr){3-4} \cmidrule(lr){5-6}
    Genre & Tasks & \vista & \ours & \vista & \ours \\
    \midrule
    \multicolumn{6}{l}{\emph{\Hrt}} \\
    \quad Shooter & 18 & 24.1 & \cellcolor{apbg}63.0 & 11.1 & \cellcolor{apbg}55.6 \\
    \quad Fighting & 11 & 6.1 & \cellcolor{apbg}72.7 & 18.2 & \cellcolor{apbg}36.4 \\
    \quad Action (other) & 11 & 15.2 & \cellcolor{apbg}51.5 & 27.3 & \cellcolor{apbg}45.5 \\
    \multicolumn{6}{l}{\emph{\Brt}} \\
    \quad Platformer & 29 & 42.5 & \cellcolor{apbg}62.1 & 37.9 & \cellcolor{apbg}48.3 \\
    \quad Racing \& other & 11 & 24.2 & \cellcolor{apbg}54.5 & 0.0 & \cellcolor{apbg}27.3 \\
    \multicolumn{6}{l}{\emph{\Tb}} \\
    \quad Puzzle & 20 & 63.3 & \cellcolor{apbg}83.3 & 45.0 & \cellcolor{apbg}70.0 \\
    \quad Card \& strategy & 9 & 55.6 & \cellcolor{apbg}92.6 & 44.4 & \cellcolor{apbg}77.8 \\
    \quad Story, sim \& other & 11 & 60.6 & \cellcolor{apbg}75.8 & 45.5 & \cellcolor{apbg}63.6 \\
    \bottomrule
  \end{tabular}
  \caption{\textbf{Success by game genre} in \% in real time at low effort, averaged over the episodes of a task and then over tasks. Genres come from a keyword rule on each game's description, and genres with fewer than six tasks join ``other''.}
  \label{tab:supp-genre}
\end{table}

\FloatBarrier
\subsection{Other models}
\label{sec:supp-models}

\paragraph{Both harnesses with four more models.}
\cref{tab:main} includes \vista and \ours with three open-source models, DeepSeek-V4.1-Flash, GLM-5.3-Flash and Hy4-flash.
It also includes one more closed model, Qwen4-Max.
We access them through OpenAI-compatible APIs and run them through the same Codex app server as \gpt.
DeepSeek-V4.1-Flash, GLM-5.3-Flash and Qwen4-Max think at low effort. Hy4-flash has a single thinking mode.
Budgets, tasks and rules are those of the main experiments, with one episode per task.
\ours wins more tasks than \vista with every model under both clocks, and every comparison in \cref{tab:supp-newmodels} is significant.
Pausing the clock helps these models much less than it helps \vista with \gpt, and only Qwen4-Max gains clearly with both harnesses.
The gap between the harnesses thus remains when the game waits, so with these models it does not come mainly from the running clock.

\paragraph{Turn-based environments.}
\ours also leads with these models in \tb environments, where latency costs nothing.
Much of the gain thus comes from executing a whole plan per model call rather than from avoiding latency.

\paragraph{GameWorld.}
On GameWorld, \ours also wins far more tasks than \vista with Qwen4-Max and Hy4-flash, as \cref{tab:supp-newmodels-gw} shows.

\paragraph{Provider limits and cost.}
These providers reject requests above a size limit.
\vista sends its visual memory and \ours its recent frames with each request, so long episodes can exceed it, and such episodes count as failures.
With Qwen4-Max and DeepSeek-V4.1-Flash, a real-time episode costs well under a dollar with either harness.

\begin{table}[!htbp]
  \centering
  \footnotesize
  \setlength{\tabcolsep}{3pt}
  \begin{tabular}{llcccc}
    \toprule
    Model & Clock & \vista & \ours & Only one wins & $p$ \\
    \midrule
    DeepSeek-V4.1-Flash & real time & 20.0 & \cellcolor{apbg}39.2 & 1 / 24 & $2\times10^{-6}$ \\
     & paused & 24.2 & \cellcolor{apbg}44.2 & 1 / 25 & $8\times10^{-7}$ \\
    Qwen4-Max & real time & 14.2 & \cellcolor{apbg}27.5 & 1 / 17 & $1.5\times10^{-4}$ \\
     & paused & 25.0 & \cellcolor{apbg}41.7 & 2 / 22 & $4\times10^{-5}$ \\
    GLM-5.3-Flash & real time & 3.3 & \cellcolor{apbg}18.3 & 0 / 18 & $8\times10^{-6}$ \\
     & paused & 1.7 & \cellcolor{apbg}26.7 & 1 / 31 & $2\times10^{-8}$ \\
    Hy4-flash & real time & 3.3 & \cellcolor{apbg}32.5 & 0 / 35 & $6\times10^{-11}$ \\
     & paused & 5.8 & \cellcolor{apbg}38.3 & 1 / 40 & $4\times10^{-11}$ \\
    \bottomrule
  \end{tabular}
  \vspace{-0.5em}
  \caption{\textbf{Paired tests for the four further models on \bench.} Each model plays one episode on each of the 120 tasks. Success rates by environment type are in \cref{tab:main}. Only one wins: tasks won only by \vista / only by \ours. $p$: two-sided exact McNemar test.}
  \label{tab:supp-newmodels}
\end{table}

\begin{table}[!htbp]
  \centering
  \footnotesize
  \setlength{\tabcolsep}{3pt}
  \begin{tabular}{lcccc}
    \toprule
    Model & \vista & \ours & Only one wins & $p$ \\
    \midrule
    \gpt & 42.9 \prog{653} & \cellcolor{apbg}80.0 \prog{900} & 0 / 63 & $2\times10^{-19}$ \\
    \opus & 41.2 \prog{623} & \cellcolor{apbg}67.6 \prog{845} & 6 / 51 & $6\times10^{-10}$ \\
    Qwen4-Max & 27.6 \prog{515} & \cellcolor{apbg}57.1 \prog{758} & 3 / 53 & $8\times10^{-13}$ \\
    Hy4-flash & 20.0 \prog{399} & \cellcolor{apbg}51.8 \prog{713} & 0 / 54 & $1\times10^{-16}$ \\
    \bottomrule
  \end{tabular}
  \vspace{-0.5em}
  \caption{\textbf{Other models on GameWorld.} Success rate (\%) with mean progress in small type, in real time with one episode on each of the 170 tasks. Only one wins: tasks won only by \vista / only by \ours.}
  \label{tab:supp-newmodels-gw}
\end{table}

\paragraph{Further models with \ours only.}
\cref{tab:supp-models} runs only \ours in real time with two more models, GPT-6 Astra~\cite{openai2026gpt6astra} and Claude Fable 5.2~\cite{anthropic2026fable52}, on all 120 tasks with one episode per task.
Both do at least as well as \gpt (\cref{tab:main}).

\begin{table}[!htbp]
  \centering
  \footnotesize
  \setlength{\tabcolsep}{3pt}
  \begin{tabular}{@{}llcccc@{}}
    \toprule
    Model & Effort & \makecell{Hostile\\real-time} & \makecell{Benign\\real-time} & \makecell{Turn-\\based} & Overall \\
    \midrule
    GPT-6 Astra & low & 70.0 \prog{822} & 62.5 \prog{781} & 87.5 \prog{934} & 73.3 \prog{846} \\
     & high & 70.0 \prog{841} & 67.5 \prog{820} & 87.5 \prog{941} & 75.0 \prog{867} \\
    \midrule
    Claude Fable 5.2 & low & 70.0 \prog{810} & 60.0 \prog{791} & 85.0 \prog{945} & 71.7 \prog{849} \\
     & high & 75.0 \prog{862} & 55.0 \prog{777} & 85.0 \prog{926} & 71.7 \prog{855} \\
    \bottomrule
  \end{tabular}
  \vspace{-0.5em}
  \caption{\textbf{Further models with \ours only.} All runs are in real time with one episode per task. Small type gives mean progress.}
  \label{tab:supp-models}
  \vspace{-1em}
\end{table}

\FloatBarrier
\subsection{Tool use, ablations and cost}
\label{sec:supp-ablations}

\paragraph{Ablations, all environment types.}
\cref{tab:cost-full} extends \cref{tab:cost} to all environment types.
In the No controller \& macro variant, every failed episode runs out of its 100 model calls, and the model takes 77 steps per episode, against 659 for the full \ours on the paired episodes.
\begin{table*}[!tp]
  \centering
  \small
  \setlength{\tabcolsep}{5pt}
  \resizebox{\linewidth}{!}{\begin{tabular}{lccccccccc}
    \toprule
    Harness & \Hrt & \Brt & \Tb & Overall & Model calls & Tokens & Cost / ep. \\
    \midrule
    Codex & 17.5 & 36.7 & 62.5 & 38.9 & 81 & 5.1M & \$1.35 \\
    \vista & 16.7 & 37.5 & 60.8 & 38.3 & 80 & 5.3M & \$1.50 \\
    \midrule
    \oursrow \ours & \best{62.5} & 60.0 & \best{83.3} & \best{68.6} & \best{43} & \best{2.8M} & \best{\$0.89} \\
    \quad No controller & 50.8 & 60.0 & \best{83.3} & 64.7 & 58 & 4.6M & \$1.31 \\
    \quad No macro & \best{62.5} & \best{61.7} & 78.3 & 67.5 & 46 & 3.1M & \$0.91 \\
    \quad No controller \& macro & 22.5 & 45.0 & 70.0 & 45.8 & 81 & 5.5M & \$1.45 \\
    \bottomrule
  \end{tabular}}
  \vspace{-0.5em}
  \caption{\textbf{Ablations and cost for all environment types.} All runs use \gpt in real time. Tokens include input, output, cached and reasoning tokens. Cost is at list price. \vista's model calls are its tool calls.}
  \label{tab:cost-full}
  \vspace{-1em}
\end{table*}

\begin{figure*}[!tp]
  \centering
  \includegraphics[width=0.8\textwidth]{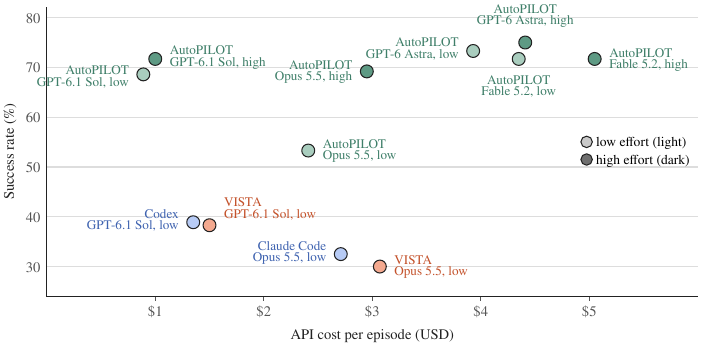}
  \vspace{-0.3em}
  \caption{\textbf{Success against cost} on \bench in real time. Color marks the harness, light and dark shades low and high effort. Cost per episode at list price. Claude Fable 5.2 cache prices assume Fable 5.1 ratios.}
  \label{fig:cost}
  \vspace{-1em}
\end{figure*}

\paragraph{Cost against success.}
\cref{fig:cost} plots success against API cost per episode for \ours with four models at low and high effort and for the baselines at low effort.
It omits the models of \cref{sec:supp-models} and \vista at high effort.

\paragraph{Cost with \opus.}
With \opus at Claude list prices, a real-time episode costs 21\% less with \ours than with \vista, as \cref{tab:cost-opus} shows.
A success costs about \$4.53 against \$10.25.
With the clock paused it costs \$4.49 against \$7.22, and \ours at high effort pays \$4.27 per success.
Part of the difference comes from outcomes, since lost episodes use more of the budget.

\begin{table}[!htbp]
  \centering
  \footnotesize
  \setlength{\tabcolsep}{3pt}
  \begin{tabular}{@{}llcccc@{}}
    \toprule
    Setting & Model & \ours & \vista & Both win & Both lose \\
    \midrule
    \bench, real time & \gpt & 0.89 & 1.50 & 0.42 / 0.68 & 1.65 / 2.01 \\
     & \opus & 2.41 & 3.07 & 0.94 / 1.27 & 3.89 / 3.80 \\
    \bench, paused & \opus & 2.73 & 2.83 & 0.86 / 1.37 & 4.88 / 3.72 \\
    GameWorld, real time & \gpt & 0.56 & 1.35 & 0.20 / 0.39 & 1.35 / 2.06 \\
     & \opus & 2.02 & 2.53 & 0.41 / 0.67 & 4.37 / 3.91 \\
    \bottomrule
  \end{tabular}
  \vspace{-0.5em}
  \caption{\textbf{Cost per episode} in \$ at list price. Both win and both lose: paired episodes that both harnesses win or both lose, given as \ours / \vista.}
  \label{tab:cost-opus}
\end{table}

\paragraph{Successful and failed episodes.}
Of the 360 real-time \ours episodes with \gpt in \cref{tab:main}, 247 succeed and 113 fail (\cref{tab:wonlost}).
Successful \ours episodes end early, with 32 model calls and 335 steps on average, against 66 model calls and 1436 steps in failed ones.
Failed episodes rely more on \tool{python}, with 35\% of calls against 20\%, and less on \tool{act}, with 34\% against 51\%.
Their controllers send 93\% of their steps against 84\%.
When a game goes badly, the model keeps rewriting its controller.
This also holds within tasks: among the 23 tasks with both successes and failures across their three episodes, the failed episodes have the higher \tool{python} share in 19.
The controller share thus reflects how the model reacts to trouble, rather than which tool wins.

\begin{table}[!htbp]
  \centering
  \footnotesize
  \setlength{\tabcolsep}{3pt}
  \begin{tabular}{lccccccc}
    \toprule
     & \multicolumn{4}{c}{Tool calls (\%)} & \multicolumn{3}{c}{Steps (\%)} \\
    \cmidrule(lr){2-5}\cmidrule(l){6-8}
    Episodes & \tool{act} & macro & \tool{python} & other & model & macro & ctrl. \\
    \midrule
    Won (247) & 51 & 24 & 20 & 5 & 5 & 11 & 84 \\
    Lost (113) & 34 & 27 & 35 & 5 & 2 & 5 & 93 \\
    \bottomrule
  \end{tabular}
  \vspace{-0.5em}
  \caption{\textbf{Tool use in successful and failed episodes} of \ours with \gpt in real time, pooled over episodes. Here ``other'' means read-only tools and ``ctrl.'' means controllers.}
  \label{tab:wonlost}
  \vspace{-1em}
\end{table}

\paragraph{Who is playing when.}
In the direct-API runs, a controller plays alone for 55.5\% of game time, the model thinks while no controller runs for 40.3\%, and both happen at once for 1.5\%, which is 3.5\% of inference time.
With the Codex app server, a controller runs during 2.3\% of inference time.
In \hrt environments the first two shares are 70.1\% and 27.4\%, and the overlap is 6.2\% of inference time.
Of the 3{,}239 automatic restarts, 74\% happen while a controller plays alone, 24\% while the model thinks, and 1\% while both are active.
A model call takes 6.1\,s at the median, 10.7\,s at the 75th percentile and 18.1\,s at the 90th.
A controller step takes 0.41\,s at the median.
The action itself takes about half of it and the model decides its duration.
The screenshot and state read that follow take the rest.
Handing off while the controller keeps playing is supported but rarely used: in 83\% of handoffs the controller stops within 1\,s.
In a variant that kept the controller running and woke the model every 120\,s, tested on 36 tasks, the share of inference time during which a controller kept playing rose from 7.8\% on these tasks to 26\% without more wins.
It won 7 of the 36 tasks, against 8 in a standard run on the same tasks.

\paragraph{Open-loop controllers.}
To test whether controllers need to read the screen while they run, we run \ours with \gpt with controllers that cannot read the screen.
The run plays one episode per task.
In it, \texttt{api.observe()} is refused, and \texttt{api.act()} reports only success and game end.
The model still sees the frames when it writes a controller, so such a controller is a timed program planned in advance, like a macro with loops and timing.
Without frames, the controller has no screen watch, and its tool description is adapted accordingly.
Macros, the model's own frames, restart notices and budgets are unchanged.
Programs that cannot read the screen at run time still win 78 of 120 tasks, against 83 for the full harness and 80 without controllers.
Exact McNemar tests give $p=0.30$ and $p=0.79$.
Against the full harness, 5 tasks are won only by the open-loop version and 10 only by the full one.
By environment type, they win 23, 25 and 30 in \hrt, \brt and \tb environments, against 26, 23 and 34 for the full harness.
Most of the gain over \vista thus comes from delegating timed action sequences, which macros already provide.
Reading the screen at run time adds five wins, a difference that is not significant.

\paragraph{Shorter context.}
To test whether \ours needs the full context history, we cap its context window at 128k tokens, so that Codex's context compaction triggers at 56k tokens.
The run uses \gpt with one episode per task on the same tasks as the first main-table run.
It wins 76 of 120 tasks against 83 for the uncapped run ($p=0.118$), with 4 tasks won only with the cap and 11 only without.
Among the 72 tasks whose context exceeded 56k tokens in the uncapped run, it wins 35 against 39 ($p=0.34$).
Neither difference is significant.

\FloatBarrier
\subsection{Failures and restarts}
\label{sec:supp-restarts}

\paragraph{How episodes end.}
In real time, all 113 failed episodes of \ours end with a used-up budget: game time in 44, steps in 42 and model calls in 27.
Of \vista's 222 failed episodes, 208 end by exhausting its 100 model calls and 14 end earlier, before any budget is used up.
\vista spends 544\,s of game time per episode on average and \ours 829\,s. With 4000\,s of game time, only 3 of \ours's successes come after 2000\,s, and its remaining failures run out of steps in 19 episodes and of model calls in 16, rather than of time.

\paragraph{Restarts.} With \gpt in real time and counting every automatic restart, \ours restarts 9.1 times per episode on average and \vista 7.9.
With \opus, they restart 9.3 and 5.1 times.
With this count, as in \cref{fig:curves}a, \ours wins 45.3\% of its 360 episodes without any restart and \vista 35.0\%.
Within 20 restarts, they reach 65.8\% and 38.1\%, and 54.2\% and 15.8\% in \hrt environments.
With \gpt, one episode per task and three times the model calls, \vista rises to 55.0\% overall and 37.5\% in \hrt environments, as \cref{tab:supp-budgets} shows.
This is still below \ours with the standard budget, at 69.2\% overall and 65.0\% in \hrt environments.
\ours itself gains little from 300 model calls and reaches 71.7\%.

\begin{figure*}[!tp]
  \centering
  \includegraphics[width=\textwidth]{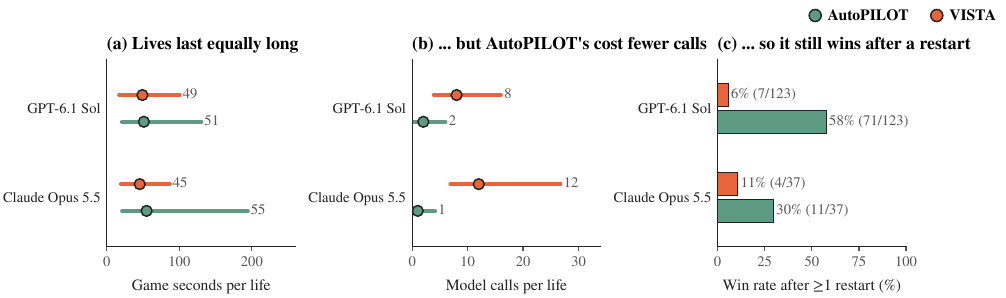}
  \caption{\textbf{Lives and restarts} in real time with \gpt and \opus. A \emph{life} is the playable game time between consecutive restarts. The last, censored segment of each episode is excluded from (a) and (b).
  (a) Game seconds and (b) model calls per life: median over episodes of each episode's median life, with interquartile range, over episodes with at least one restart.
  (c) Success on paired episodes in which both harnesses restarted: \gpt 71/123 \vs 7/123 ($p\approx10^{-19}$) and \opus 11/37 \vs 4/37 ($p=0.039$).}
  \label{fig:lives}
\end{figure*}

\paragraph{Lives.}
\cref{fig:lives} splits episodes into lives at each restart.
A life lasts about as long under both harnesses.
Its median is 51 \vs 49\,s with \gpt and 55 \vs 45\,s with \opus.
However, an \ours life costs 2 model calls against 8 for \vista with \gpt, and 1 against 12 with \opus.
After a failure, \vista thus has fewer model calls left for the next attempt.
On the paired episodes of \cref{fig:lives}c, it used up its 100-call budget in 110 of 123 with \gpt and in 29 of 37 with \opus, against 5 and 4 for \ours.
So after an automatic restart, \ours still wins 58\% of paired episodes against 6\% for \vista with \gpt, and 30\% against 11\% with \opus.

\begin{figure*}[!tp]
  \centering
  \includegraphics[width=0.78\textwidth]{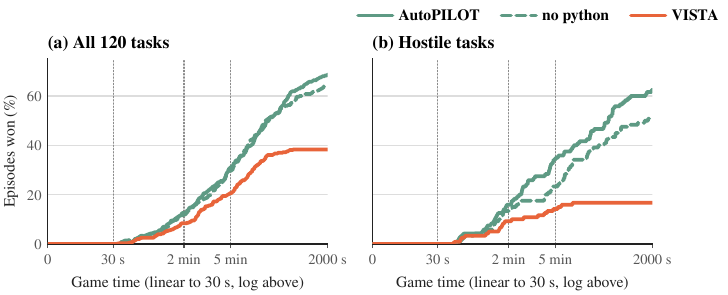}
  \caption{\textbf{Success over game time} with \gpt at low effort in real time. Panel (b) shows \hrt only. Share of all episodes won by game time $t$. Failures stay in the denominator, so the right end equals the success rate. Dotted lines mark 30\,s, 2\,min and 5\,min.}
  \label{fig:time-to-win}
\end{figure*}

\paragraph{Success over game time.}
\cref{fig:time-to-win} shows the share of all launched episodes that have succeeded by each moment of game time, with \gpt at low effort in real time.
Failed episodes stay in the denominator.
No harness wins within 30\,s, and the earliest win comes after 32\,s.
At 2 and 5\,min, \ours has won 12.5\% and 30.8\% of all episodes, \ours without controllers, labeled no python in the figure, 11.9\% and 29.7\%, and \vista 8.3\% and 20.6\%.
In \hrt environments, the numbers are 15.8\% and 34.2\%, 13.3\% and 23.3\%, and 9.2\% and 14.2\%.
The gap opens late rather than early: \ours leads \vista by 4.2 points at 2\,min, 10.2 at 5\,min and 30.3 at 2000\,s.
In \hrt environments the lead is 6.6, 20.0 and 45.8 points.
This is because \ours keeps winning after \vista has run out of model calls.
\vista's last win comes at 1014\,s, and at 419\,s in \hrt environments.

\FloatBarrier
\subsection{Scaling with budget, effort and speed}
\label{sec:supp-scaling}

\paragraph{Budget.}
\cref{tab:supp-budgets} breaks down the budget variants of \cref{tab:budgets-main} by environment type.
Relaxing one limit at a time changes little for \ours.
Relaxing all three together helps, mostly in \hrt environments, and the gain is significant with exact McNemar $p=0.007$.
\vista gains from more model calls but not from more time or steps.
Under the relaxed budget \ours stays far ahead, so the gap between the harnesses does not come from the budget.

\begin{table*}[!tp]
  \centering
  \footnotesize
  \setlength{\tabcolsep}{3pt}
  \begin{tabular}{@{}lccccc@{}}
    \toprule
    Budget & Limits & \Hrt & \Brt & \Tb & Overall \\
    \midrule
    Standard & 100 / 2000 / 2000\,s & 65.0 \prog{806} & 57.5 \prog{761} & 85.0 \prog{918} & 69.2 \prog{828} \\
    3$\times$ model calls & 300 / 2000 / 2000\,s & 60.0 \prog{757} & 70.0 \prog{807} & 85.0 \prog{924} & 71.7 \prog{829} \\
    2$\times$ steps & 100 / 4000 / 2000\,s & 67.5 \prog{810} & 62.5 \prog{801} & 82.5 \prog{907} & 70.8 \prog{839} \\
    2$\times$ game time & 100 / 2000 / 4000\,s & 65.0 \prog{812} & 60.0 \prog{772} & 87.5 \prog{919} & 70.8 \prog{834} \\
    Relaxed & 300 / 4000 / 4000\,s & 77.5 \prog{871} & 70.0 \prog{846} & 87.5 \prog{942} & 78.3 \prog{886} \\
    \midrule
    \multicolumn{6}{@{}l}{\emph{\vista}} \\
    Standard & 100 / --$^\S$ / 2000\,s & 17.5 \prog{429} & 32.5 \prog{576} & 62.5 \prog{777} & 37.5 \prog{594} \\
    3$\times$ model calls & 300 / --$^\S$ / 2000\,s & 37.5 \prog{557} & 50.0 \prog{668} & 77.5 \prog{905} & 55.0 \prog{710} \\
    Relaxed & 300 / --$^\S$ / 4000\,s & 30.0 \prog{535} & 45.0 \prog{673} & 80.0 \prog{918} & 51.7 \prog{708} \\
    \bottomrule
  \end{tabular}
  \caption{\textbf{Scaling the budget, by environment type} with \gpt in real time on 120 tasks, with mean progress in small type. Limits are model calls / steps / game time. $^\S$Not applicable to \vista, which sends one step per model call.}
  \label{tab:supp-budgets}
\end{table*}

\paragraph{Reasoning effort and model interface.}
Calling \gpt through a direct API loop instead of the Codex app server changes success little, as \cref{tab:supp-seeds} shows.
Higher reasoning effort helps \opus clearly and \gpt only a little.
With \opus, high effort wins 83 instead of 64 of the 120 tasks, with exact McNemar $p=5\times10^{-4}$, and roughly doubles the output of each model call.
Claude Fable 5.2 wins the same number of tasks at both efforts, as \cref{tab:supp-models} shows.
With the clock paused and high effort, \gpt wins 92 of 120 tasks and \opus 84.

\begin{table*}[!tp]
  \centering
  \small
  \setlength{\tabcolsep}{7pt}
  \begin{tabular}{lcccccc}
    \toprule
     & \multicolumn{5}{c}{Generation speed (tokens/s)} & \\
    \cmidrule(lr){2-6}
    Model & 10 & 25 & 50 & 100 & 200 & Paused \\
    \midrule
    \gpt, high & 55.0 \prog{738} & 61.3 \prog{762} & 68.8 \prog{819} & 66.3 \prog{809} & 66.3 \prog{805} & 71.3 \prog{846} \\
    \gpt, low & 55.0 \prog{760} & 56.3 \prog{762} & 61.3 \prog{782} & 63.8 \prog{798} & 63.8 \prog{799} & 66.3 \prog{821} \\
    \opus, high & 31.3 \prog{522} & 42.5 \prog{658} & 46.3 \prog{658} & 58.8 \prog{774} & 62.5 \prog{770} & 63.8 \prog{810} \\
    \opus, low & 36.3 \prog{604} & 43.8 \prog{622} & 45.0 \prog{661} & 50.0 \prog{703} & 48.8 \prog{677} & 52.5 \prog{729} \\
    \bottomrule
  \end{tabular}
  \vspace{-0.5em}
  \caption{\textbf{Success rate at controlled generation speed} for \ours on 80 \hrt and \brt tasks, with mean progress in small type. The game is frozen during generation and then runs for $N/v$\,s for $N$ output tokens.}
  \label{tab:speed}
  \vspace{-1em}
\end{table*}

\paragraph{Generation speed.}
The study of \cref{fig:curves}c controls latency directly, and \cref{tab:speed} gives its numbers.
The game is frozen while the model generates and then runs for $L=N/v$ seconds for $N$ output tokens before the model call takes effect, with no fixed overhead.
The study uses the 80 \brt and \hrt tasks and the standard budget.
Each episode is checked for timing errors and repeated if a check fails.
Success rises with generation speed and levels off above about 100 tokens/s.
The trend is clear within tasks, with $p=0.005$ in a conditional logistic regression with tasks as strata, and weaker when pooled over tasks.
Without synthetic latency the same tasks give about the success of the 50 tokens/s setting, and pausing the clock adds only a little more.
Faster models will thus narrow but not remove the cost of thinking in real time, and \ours already recovers most of what pausing the clock would give.

\paragraph{Generation speed, other models.}
\cref{tab:speed} also covers high effort and \opus, and all four settings win more with faster generation.
The latency of a model call counts all its output tokens, including reasoning, so a model that writes more per call waits longer at the same speed.
\opus writes about twice as many tokens per call as \gpt, yet at matched waiting time \gpt still wins more.
At 10 tokens/s, \opus loses almost all the tasks that only \gpt wins because game time runs out.

\FloatBarrier
\subsection{GameWorld}
\label{sec:supp-gameworld}

With \gpt and the clock paused, \vista succeeds in 84.1\% of GameWorld tasks and \ours in 82.9\%.
\vista alone wins 9 tasks and \ours alone 7, as \cref{tab:supp-ci} shows.
With \opus, \vista succeeds in 67.1\% with the clock paused and 41.2\% in real time, and \ours in 72.4\% and 67.6\%.
Running the clock thus costs \vista 25.9 points and \ours 4.7.
\cref{tab:supp-newmodels-gw} adds two more models from \cref{sec:supp-models} in real time.

GameWorld tasks are shorter than those of \bench.
With the clock paused, \vista wins them with a median of 28 model calls against 45 on \bench.
Of its GameWorld wins, 55\% take at most 30 model calls, against 33\% on \bench.
Under the same clock, \ours wins them with a median of 18 model calls and 55 steps, against 30 and 117 on \bench.
Of its GameWorld wins, 66\% take at most 100 steps, against 46\% on \bench.

In 60\,s trials, 34 of the 170 GameWorld tasks can also be completed by random actions or by doing nothing.
We keep them for comparability with prior work.

\FloatBarrier
\section{Case studies}
\label{sec:supp-cases}
\begin{figure*}[p]
  \centering
  \includegraphics[width=0.93\textwidth]{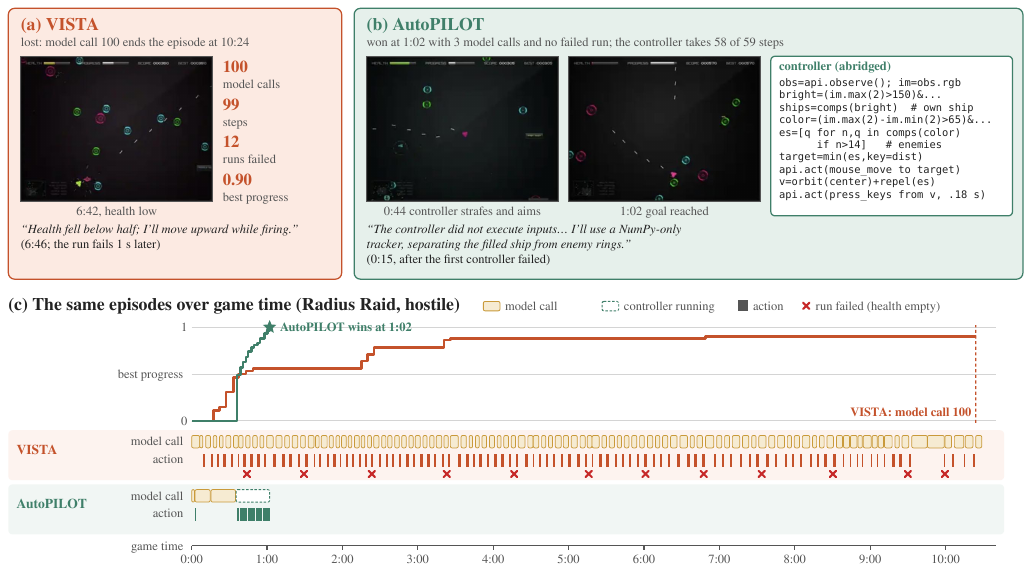}
  \caption{\textbf{Case study: Radius Raid}, a \hrt game, with \gpt in real time. (a) \vista at 6:42, shortly before its eighth failure. (b) An abridged version of \ours's controller, which tracks the ship and enemies by color. (c) Best progress, model calls, steps and failures over game time. \vista loses its progress at every failure, while \ours wins at 1:02.}
  \label{fig:case-radius}
\end{figure*}

\begin{figure*}[p]
  \centering
  \includegraphics[width=0.93\textwidth]{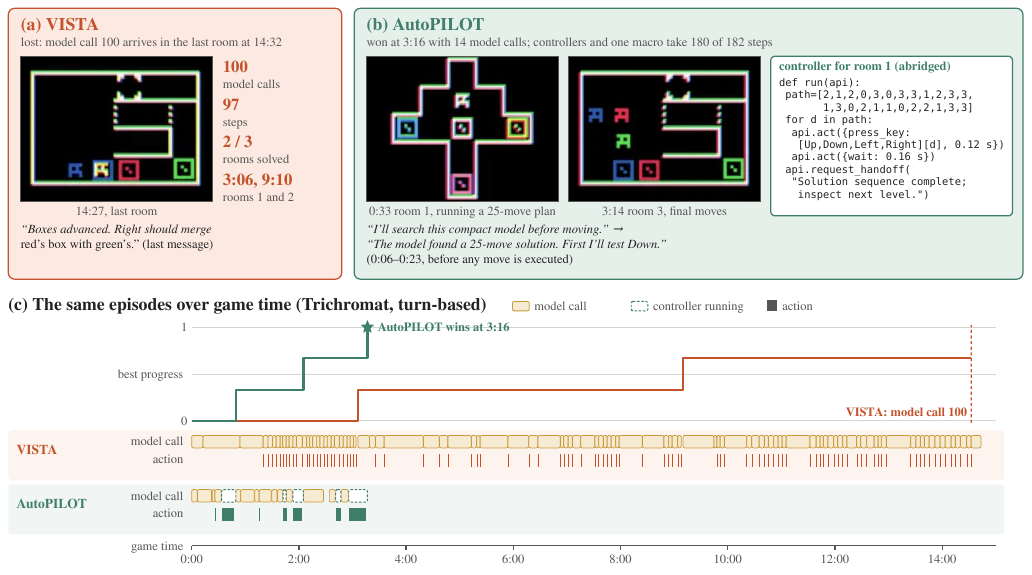}
  \caption{\textbf{Case study: Trichromat}, a \tb game, with \gpt in real time. (a) \vista moves once per model call and reaches the last room with its last calls. (b) \ours searches a Python model of each room and executes the solution with controllers, shown abridged for room 1. (c) Best progress, model calls, controller runs and steps over game time. Each room adds a third to progress.}
  \label{fig:case-trichromat}
\end{figure*}

\paragraph{Recorded episodes.}
\cref{tab:cases} summarizes three recorded real-time episodes with \gpt.
\cref{fig:case-radius,fig:case-trichromat} replay two of them in detail, one for each way a single-step harness fails.
Frames, quotes, timings and code come from the recorded model outputs and action logs.

\begin{table}[!htbp]
  \centering
  \footnotesize
  \setlength{\tabcolsep}{2.5pt}
  \begin{tabular}{@{}p{0.15\linewidth}p{0.4\linewidth}p{0.4\linewidth}@{}}
    \toprule
    Game & \vista & \ours \\
    \midrule
    PRAXIS FIGHTER X \newline {\scriptsize\hrt} &
    One action per model call, each after several seconds of thinking. The ship is often destroyed before the action lands. 30 failures in 100 tool calls. Best progress 25\%. \textbf{Lost.} &
    A macro ends with the first failure at 0:14, and two more follow by 0:44. At 0:32 the model writes the screen-driven controller of \cref{lst:controller}b. It takes over at 0:44, plays 103 steps without another failure and wins at 1:26, after 6 model calls in total. \textbf{Won.} \\
    \addlinespace
    Radius Raid \newline {\scriptsize\hrt} &
    One strafe per model call. Health runs out 12 times. Model call 100 ends the episode at 10:24. \textbf{Lost.} &
    One \tool{act} turns on autofire. Two model calls write a targeting controller that takes 58 of 59 steps. Won at 1:02 with 3 model calls, no loss. \textbf{Won.} \\
    \addlinespace
    Trichromat \newline {\scriptsize\tb} &
    One tile per model call. Solves two of three rooms at 3:06 and 9:10. Model call 100 arrives in the last room at 14:32. \textbf{Lost.} &
    Writes each room as a Python model and searches it, then reports ``found a 25-move solution''. Tests one move, then executes the rest with controllers and one macro. 180 of 182 steps come from the worker. Won at 3:16 with 14 model calls. \textbf{Won.} \\
    \bottomrule
  \end{tabular}
  \vspace{-0.5em}
  \caption{\textbf{Three recorded episodes} with \gpt in real time. Times are game time. Recorded separately from the main-table runs.}
  \label{tab:cases}
  \vspace{-1em}
\end{table}

\paragraph{Radius Raid: enough skill, not enough time.}
In this \hrt game, \vista strafes once per model call and reaches 88\% of the goal by 3:26, as \cref{fig:case-radius} shows.
But enemies drain its health while it thinks, and it fails 12 times.
Every restart resets progress, so its 100 model calls run out at 10:24 with a best progress of 90\%.
\ours turns on autofire with one \tool{act}, and its first controller fails to send actions.
The model reads the failure from the next frame and writes a second controller that segments the ship and the enemies by color, aims at the nearest enemy and orbits the arena while avoiding contact.
Then the model waits.
The controller takes 58 of the episode's 59 steps and reaches the goal at 1:02, after three model calls and without a failure.

\paragraph{Trichromat: enough time, not enough model calls.}
Waiting costs nothing in this turn-based puzzle, so latency is not the problem.
The problem is the model-call budget, as \cref{fig:case-trichromat} shows.
\vista spends one model call per move and solves two of the three rooms, at 3:06 and 9:10, before its 100th call arrives in the last room.
\ours writes the first room as a Python model and searches it before moving, tests the first move of the 25-move solution with \tool{act}, and executes the rest with one controller.
When its model of the second room has no solution, it tests the unknown mechanic, a floor button that opens a door, revises the model, and again executes the plan in bulk.
It solves all three rooms by 3:16 with 14 model calls.
Controllers and one macro take 180 of its 182 steps.

\FloatBarrier
\section{Limitations}
\label{sec:limits}

\ours relies on model-written perception and control code, which is easier in 2D browser games than in cluttered 3D scenes.
In \hrt environments most of its wins come after a restart, so winning an unseen real-time game at the first attempt remains open.
Following Active-Prompt~\cite{diao-etal-2024-active}, future work could collect human demonstrations from other games at points where the model is uncertain.
Models also rarely let a controller keep playing after a handoff, and doing so by default did not add wins, as \cref{sec:supp-ablations} reports.

\end{document}